\documentclass[journal,10 pt,twocolumn]{IEEEtran}

\usepackage{graphicx}
\usepackage{balance}
\usepackage{comment}
\usepackage{cite}
\usepackage{amssymb}
\usepackage[tight,footnotesize]{subfigure}
\usepackage[active]{srcltx}
\usepackage{amsmath}
\usepackage{mathtools}

\graphicspath{{./figs/}}
\renewcommand{\baselinestretch}{0.95}
\usepackage{eurosym}
\usepackage[plain]{algorithm}
\usepackage{algorithmic}
\usepackage{multicol}
\usepackage{dsfont}
\usepackage{amsfonts}

\usepackage{cases}
\usepackage{xcolor}
\makeatletter
\renewcommand{\ALG@name}{Program}
\makeatother

\newlength\mylength 

\begin{document}

\title{Towards Automated 3D Search Planning for Emergency Response Missions}

\author{Savvas~Papaioannou,~Panayiotis~Kolios,~Theocharis~Theocharides,\\~Christos~G.~Panayiotou~ and ~Marios~M.~Polycarpou% <-this % stops a space
\thanks{The authors are with the KIOS Research and Innovation Centre of Excellence (KIOS CoE) and the Department of Electrical and Computer Engineering, University of Cyprus, Nicosia, 1678, Cyprus. {\tt\small \{papaioannou.savvas, pkolios, ttheocharides, christosp, mpolycar\}@ucy.ac.cy}}
}

\markboth{Journal of Intelligent \& Robotic Systems 103, no. 1 (2021): 2, doi:10.1007/s10846-021-01449-4}%
{Papaioannou \MakeLowercase{\textit{et al.}}: Towards automated 3d search planning for emergency response missions}

\maketitle

\begin{abstract}
The ability to efficiently plan and execute automated and precise search missions using unmanned aerial vehicles (UAVs) during emergency response situations is imperative. Precise navigation between obstacles and time-efficient searching of 3D structures and buildings are essential for locating survivors and people in need in emergency response missions. In this work we address this challenging problem by proposing a unified search planning framework that automates the process of UAV-based search planning in 3D environments. Specifically, we propose a novel search planning framework which enables automated planning and execution of collision-free search trajectories in 3D by taking into account low-level mission constrains (e.g., the UAV dynamical and sensing model), mission objectives (e.g., the mission execution time and the UAV energy efficiency) and user-defined mission specifications (e.g., the 3D structures to be searched and minimum detection probability constraints). The capabilities and performance of the proposed approach are demonstrated through extensive simulated 3D search scenarios.
\end{abstract}


\section{Introduction} \label{sec:Introduction}

The miniaturization and cost reduction of electronic components and the recent technological advancements in avionics, robotic systems and artificial intelligence has led to the rapid growth of unmanned aerial vehicles (UAVs). We have now reached a state where UAVs have become an important asset in various application domains including response efforts in disaster management \cite{Tomic2012,Goodrich2008,Bitton2008,Erdelj2016,Papaioannou2019_1,Polka2017,Silvagni2017,Scherer2015,Erdos2013,Papaioannou2019_2,Papaioannou2020,Papaioannou2021,Ra,Rb,Rc,Rd}. For instance, the use of UAVs in search and rescue (SAR) missions not only can allow for more efficient organization, planning and execution of tasks, but it can also enhance the safety of the first responders by allowing them a) to analyze the situation at hand before proceeding with their operations and b) assisting them during their search operations by providing information about their searching patterns and spreading.

\begin{figure}
	\centering
	\includegraphics[scale=0.25]{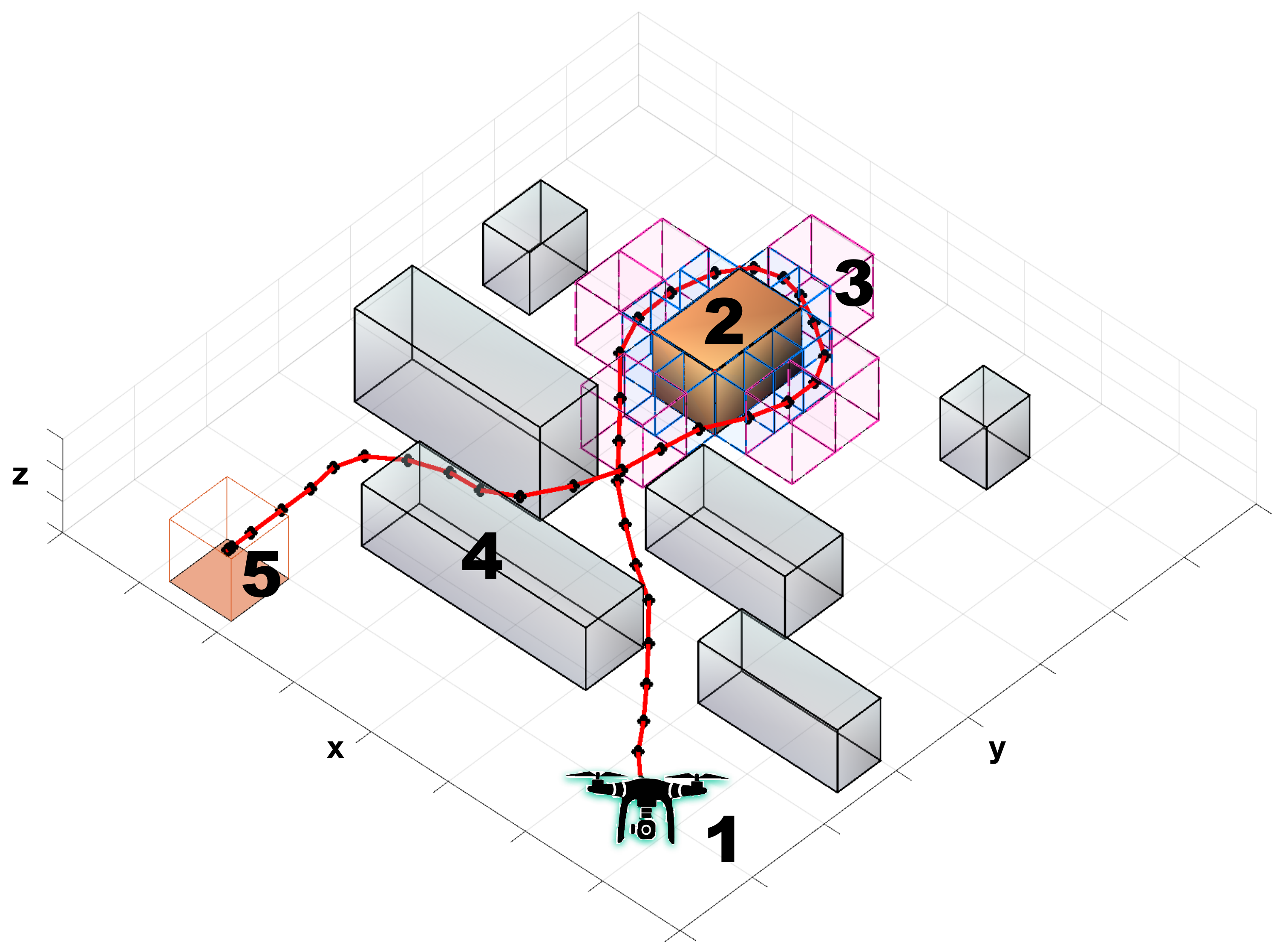}
	\caption{The paper proposes a search-planning framework which can be used by first-responders in order to automate the UAV-based 3D searching during SAR missions. The proposed framework allows the UAV agent to search an object of interest in 3D, avoid obstacles and optimize the mission objectives. The numbers indicate: 1) UAV starting position, 2) the object of interest to be searched, 3) cuboids generated by the proposed framework which allow the UAV to search the object of interest in 3D, 4) obstacles, 5) goal region to be reached by the UAV at the end of the mission. Finally, the red line shows the generated planned UAV trajectory.}	
	\label{fig:scenario}
	\vspace{-0mm}
\end{figure}

However, nowadays UAVs are being used by first responders mainly to provide a birds-eye view of the incident (in case of a forest fire or flood, for instance) and for conducting rapid spot searches over inaccessible areas (e.g., to locate missing people and for damage assessment). In many situations, searching the affected area for survivors from a birds-eye view is not sufficient for locating survivors, especially in challenging environments (e.g., under foliage). In addition, precise navigation between obstacles and time-efficient searching of 3D structures are essential for the success of the mission \cite{IFAFRI2}. Moreover, UAVs are currently being operated manually which can be an inefficient and error-prone process. Finally, the manual operation of UAVs requires a high degree of human expertise and substantial training, and as a result the operation of UAVs is limited only to specialists \cite{USAR}.

To summarize, when natural disasters occur an immediate life-saving response is essential in order to rescue people from imminent danger. The goal of this life-saving response i.e., the search and rescue (SAR) mission is to rescue the largest number of people in the shortest time, while at the same time minimize the risk of the rescuers. Search missions however, could be extremely challenging and dangerous. The search crew is required to respond to devastations caused by floods, storms, maritime accidents, earthquakes, hazardous materials releases, etc. The responders are often required to posses specialized skills, training and equipment in order to work in areas where public services are unavailable and the infrastructure is destroyed and disrupted (e.g., during floods with downed power lines and gas leaks). The search team is often required to search around and along large structures/buildings, below bridges and under high foliage in order to locate people in need. The fundamental idea which motivates this work, is that an autonomous aerial agent (i.e., a UAV) could become an important aid in many search and rescue missions by improving the efficiency and organization of the mission while at the same time reducing the need to place the rescuers in danger situations. However, UAVs are currently under-utilized in SAR missions as they are mainly used to provide an aerial birds-eye view of the situation at hand. For this reason, in this work we propose a novel search-planning framework which can be used by first responders in order to automate the operation of an autonomous UAV agent in search-missions, and allow for precise, efficient and collision-free UAV-based navigation and searching around 3D structures.

More specifically, this work proposes a search planning framework that can be used by the rescue crew in order to automate the planning and execution of UAV-based search missions in 3D environments. An illustrative example is shown in Fig. \ref{fig:scenario}. The proposed framework takes into account a) the low-level mission constraints such as the UAV dynamics (we model the UAV agent as a point mass with 3D linear dynamics) and sensing model (i.e., the geometry of the sensor's field-of-view and sensing constraints), b) the mission objectives (i.e., the mission execution time and the UAV energy efficiency) and finally c) the search mission specifications (e.g., the objects to be searched, the obstacles to be avoided, and the probability of detecting survivors) and then it provides a collision-free search trajectory which optimizes the mission objectives and searches the objects of interests in 3D with the required user specified detection probability. Overall, the contributions of this work are as follows:

\begin{itemize}
	\item We propose a novel search planning framework which can be used to automate the UAV-based search missions in 3D environments. The proposed framework takes into account low-level mission constrains (UAV dynamical and sensing model), mission objectives (mission execution time and the UAV energy efficiency) and mission specifications (e.g., 3D structures to be searched, obstacles to be avoided and the minimum detection probability for locating survivors) and computes the optimal UAV control inputs that execute the desired search plan. 
	\item We model all 3D structures (i.e., objects to be searched and obstacles) using rectangular cuboid primitives which allows us to use mathematical programming techniques in order to formulate the search planning problem in 3D using mixed integer quadratic programming (MIQP) and solve it exactly using standard off-the-shelf solvers. 
\end{itemize}

\noindent To briefly summarize the proposed approach, in this work we assume that a controllable UAV agent, operates inside a bounded 3D environment. The UAV agent evolves in 3D according to its dynamical model (i.e., Sec. \ref{ssec:agent_dynamics}) and can search the surrounding area for survivors according to its sensing model (i.e., Sec. \ref{ssec:sensing_model}). The environment contains two types of objects i.e., a) \textit{objects of interest} which need to be searched in 3D (i.e., searched across all faces), and b) \textit{obstacles} that need to be avoided. All types of objects are modeled as rectangular cuboids in this work (i.e., Sec. \ref{ssec:cuboids}). The goal of the UAV agent is to search all faces of each object of interest for survivors with the required user specified minimum probability of detection and navigate to a user specified goal region by avoiding the obstacles in its way. At the same time the UAV agent must optimize the mission objectives i.e., mission execution time and energy efficiency. To accomplish this task in the proposed framework we use mathematical programming techniques to encode the various requirements of the mission including the description of objects of interests, obstacle avoidance constraints and the mission objectives as a mixed integer quadratic program (MIQP) which is solved exactly using off-the-shelf solvers. To accomplish the 3D search of an object of interest we discretize the area around the object by creating 3D zones (i.e., Sec. \ref{ssec:3dzone}) and with mathematical programming techniques we create 3D search constraints (i.e., Sec. \ref{ssec:3D_con}) in order to guide the UAV through the appropriate 3D zones and search the object of interest according to the required detection probability.

The rest of the paper is structured as follows. Section~\ref{sec:Related_Work} presents an overview of the related work on this topic. Section~\ref{sec:problem} formulates the problem and provides an overview of the proposed framework. Section \ref{sec:system_model} develops the system model and Sec. \ref{sec:search_planning} discusses the details of the proposed 3D search planning approach. Finally, Sec. \ref{sec:Evaluation} evaluates the proposed framework and Sec. \ref{sec:conclusion} concludes the paper and discusses future work.

\section{Related Work}\label{sec:Related_Work}

Autonomous planning and control are two of the most desirable capabilities in mobile robotics. Over the last years a plethora of methods have been proposed from academic and industrial research labs especially for the problem of autonomous planning and control for ground robots operating in 2D and 2.5D environments. For instance the authors in \cite{Colas2013} develop a path planning and execution method, for search and rescue ground robots, which is able to handle a complex and non-flat terrain. In order to reduce the computational complexity of the task, the authors propose to decouple the problem into positioning and orientation planning. Additionally, the work in \cite{Berger2015} proposes a path-planning technique for detecting a static target during search and rescue missions. The technique in \cite{Berger2015} is exact and is solved using mixed-integer linear programming (MILP). However, it is based on a 2D discrete representation of the world and it does not considers the agent dynamics. The authors in \cite{san2018} investigate various 2D path-planning heuristic techniques for searching survivors during disasters, including artificial potential fields (APF) \cite{Warren1989}, fuzzy logic \cite{Mac2016} and genetic algorithms (GA) \cite{Tu2003}. The proposed techniques however are purely kinematic and they do not consider obstacle avoidance constraints.  Similarly, the work in \cite{PapaioannouCDC19} develops a search-and-track (SAT) planning technique for searching an area of interest with multiple UAV agents and tracking multiple targets in SAR missions. Their search planning algorithm is based on integer linear programming (ILP) on a 2D discrete representation of the world without obstacles. More recently the work in \cite{Alcantara2019} investigated the problem of UAV path planning during SAR missions. The authors formulate the trajectory planning problem as a model predictive control (MPC) problem and they solve it using particle swarm optimization (PSO). This technique however used a 2D coordinated kinematic model for the UAVs and the path planning was conducted in two-dimensions.

Although the majority of proposed approaches have reached a significant level of maturity, there are still challenges to be tackled when more complex scenarios are considered i.e., autonomous UAV-based planning and control in 3D environments, complex mission objectives and low-level mission constraints. To address such challenges, the works in \cite{Bortoff2000,He2008,Tisdale2009,Yang2008} have investigated the problem of UAV-based 3D path planning with the main objective being the search for a collision-free trajectory to the goal region. Specifically, the authors in \cite{Bortoff2000} propose a two-step planning approach using Voronoi graph search and artificial forces whereas in \cite{He2008}, the  problem of UAV path planning in GPS-denied environments is being investigated using Belief Roadmaps \cite{Prentice2009}. In \cite{Tisdale2009}, a receding horizon UAV planning approach is proposed which is solved using gradient-based methods, whereas in \cite{Yang2008} the authors use rapidly-exploring random trees (RTTs) \cite{Lavalle2001} to generate collision-free waypoints which are then connected with straight line segments and smoothed out using cubic Bezier curves \cite{Han2009} to create a continuous curvature path which the UAV can execute. 

The combined problem of path and task planning, along with the automated generation and execution of high-level missions is essential for allowing autonomous robotic systems in taking part in various missions. Towards this direction, the authors in \cite{Plaku1} seek to enhance the autonomy of an under-water vehicle, by specifying the mission objectives and constraints in a high-level form using a regular language. This high-level mission specification is then automatically translated into a collision-free, dynamically feasible, and low-cost trajectory which the vehicle can execute. The approach utilizes a navigation roadmap and sampling-based motion planning to determine the dynamically feasible and collision-free trajectories along the navigation routes. In a similar fashion, the work in \cite{Plaku3}, tackles the combined problem of robot task planning and motion planning and proposes an interactive search approach, which couples sampling-based motion planning with action planning. In \cite{Plaku3}, a planning domain definition language (PDDL) is used to specify the desired robot task, which is then automatically converted into a collision-free and dynamically feasible trajectory using sampling-based motion planning. Finally, the work in \cite{Plaku2}, proposes a multi-UAV reactive motion-planner for the task of persistent coverage of risk-sensitive areas. The work in \cite{Plaku2} combines persistent coverage with risk minimization, and sensor data quality maximization, by leveraging simple interactions between the UAVs.

In this work we propose a framework dedicated to search-planning which can be used by a human operator in order to automate the UAV-based 3D trajectory planning during search missions. This is complementary to the above mentioned approaches and can be used in combination to achieve specific goals. In particular, the proposed framework is unique in the sense that it takes into account UAV characteristics, mission-specific objectives, and mission constraints and produces optimized UAV controls that allow the agent to autonomously navigate the 3D environments, avoid obstacles and search the specified 3D structures to achieve the mission objectives. While existing approaches focus mainly on avoiding obstacles while en route to their destination, the proposed framework incorporates the UAV dynamics, the UAV sensing model, and mission specific objectives into the search planning problem, thus ultimately computing the UAV control inputs which produce successful search trajectories. In addition, the proposed framework captures actual requirements found in real search-missions including: a) searching 3D objects/structures from all views and b) meeting a certain detection level in the captured data in order to spot the survivors. Finally, in the proposed framework the 3D search-planning problem is formulated as a mixed integer quadratic programming (MIQP) problem, which can be solved exactly and efficiently using off-the-shelf solvers.

\section{Problem Setup}\label{sec:problem}

\begin{figure*}
	\centering
	\includegraphics[width=\textwidth]{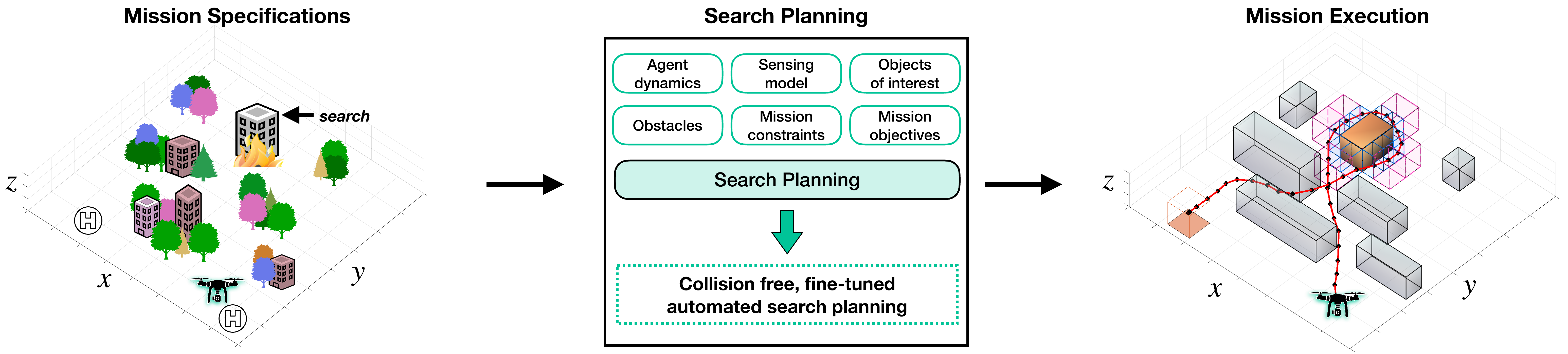}
	\caption{Overview of the proposed search-planning framework. The proposed framework aims to automate the second and most critical phase (i.e., the search phase) of a traditional SAR mission with a single UAV agent.}	
	\label{fig:arch}
\end{figure*}

SAR missions typically consist of three phases \cite{UNCHR1,UNCHR2}, namely: a) assessment, b) search and c) rescue. The goal of the assessment phase is the determination of the course of action. In this phase the rescue team assesses the damages and the hazards in the vicinity of the affected area in order to prepare and organize the search and rescue mission. Thereafter, the goal of the search phase is to conduct efficient, organized and thorough searches in the affected area in order to locate survivors as efficiently as possible. Search operations, when possible, follow optimized search patterns that have been planned ahead in order to increase the efficiency of the search. In addition, the search team is often required to search around and along large structures/buildings, below bridges and under high foliage in order to locate people in need. Finally, during the last phase, the located people are given medical aid and are transported to safety. 

With the proposed search-planning framework we aim to automate the second phase (i.e., the search phase) of the SAR mission. The proposed search planning framework is illustrated in Fig. \ref{fig:arch}. Essentially, the human operator provides the mission specifications which are then used by the proposed framework in order to automate the 3D search phase by a single UAV agent. For the \textit{Search Planning} phase, the proposed framework provides fine-tuned and collision-free search plans which incorporate information regarding the agent dynamic and sensing model, the mission objectives and constraints and the specifications regarding the objects to be searched and the obstacles to be avoided. In essence, the \textit{Search Planning} phase computes the optimal low-level control inputs which allow the UAV to autonomously search objects of interest in 3D using its camera system, avoid all obstacles in its way and navigate the surveillance region in a way which is optimal according to the specified mission objectives. 

To do so, in this work we assume that a controllable UAV agent, can operate inside a bounded 3D environment which may contain a) \textit{objects of interest} which need to be searched in 3D (i.e., searched across all faces), and b) \textit{obstacles} that need to be avoided. Additionally, we assume that the UAV departs from its home depot in the beginning of the search mission and reaches the goal region at the end of the mission. 

Let the set of all objects of interest inside the surveillance region be denoted by $J=\{j_1,j_2,\cdots,j_{|J|}\}$ with the set cardinality $|J|$ denoting their total count. Similarly, we denote the set of all obstacles in the environment by $\Xi=\{\xi_1,\xi_2,\cdots,\xi_{|\Xi|}\}$. The sets $J$ and $\Xi$ are assumed to be known and given. Additionally we assume that a) the UAV agent evolves in time according to a discrete-time dynamical model, b) the UAV is equipped with an onboard camera system which exhibits a finite field-of-view (FoV), and that c) the UAV uses its camera to take snapshots from the environment in order to find people that need help. The dimensions and location of the obstacles and objects of interest are assumed to be known.

The proposed search planning framework takes into account the low-level mission constraints i.e., the UAV dynamics and the UAV sensing model (i.e., the onboard camera system characteristics) and allows a human operator to specify the initial state (i.e., location) $x_0$ of the agent, the goal region $\mathcal{G}$ to be reached at the end of the mission (e.g., landing area), the set of objects of interest $J$ to be searched, the set of obstacles $\Xi$ to be avoided, the mission objectives (i.e., optimize the mission's execution time and/or the energy efficiency of the UAV) and finally the detection probability to be maintained during the mission i.e., the generated UAV search trajectory is computed in such a way so that the probability of detecting survivors during the mission is maintained within the required user specified level.

Finally, the above specifications are transformed using mathematical programming techniques into a mixed integer quadratic program (MIQP) which is solved exactly using standard off-the-shelf solvers. The output of the proposed framework is an optimal search plan (i.e., a sequence of low-level UAV control inputs over a finite horizon $T$ which meet the mission requirements and optimize the mission objectives).

\section{System Model} \label{sec:system_model}

\subsection{Agent Dynamics} \label{ssec:agent_dynamics}
In this work we assume that the UAV agent evolves in 3D space according to the following discrete-time linear dynamical model:
\begin{equation} \label{eq:agent_dynamics}
    x_t = \Phi x_{t-1} + \Gamma [u_{t-1} - u_g]
\end{equation}
where $x_t = [\text{x},\dot{\text{x}}]_t^\top \in \mathbb{R}^6$ denotes the agent's state at time $t$ which consists of position $\text{x}_t=[p_x, p_y, p_z]_t \in \mathbb{R}^3$ and velocity $\dot{\text{x}}_t = [\nu_x,\nu_y,\nu_z]_t \in \mathbb{R}^3$ components in 3D cartesian coordinates. The agent can be controlled by applying an amount of force $u_t  \in \mathbb{R}$ in each dimension, thus $u_{t} = [\text{u}_x, \text{u}_y, \text{u}_z]_{t}^\top$ denotes the applied force vector at $t$ and the constant $u_g = [0, 0, mg]^\top$ denotes the force of gravity where $g = 9.81 \text{m}/\text{s}^2$ is the Earth's gravitational acceleration and $m$ is the agent mass. The matrices $\Phi$ and $\Gamma$ are given by:
\begin{equation}
\Phi = 
\begin{bmatrix}
    \text{I}_{3\times3} & \Delta T \cdot \text{I}_{3\times3}\\
    \text{0}_{3\times3} & \phi \cdot \text{I}_{3\times3}
   \end{bmatrix},~
\Gamma = 
\begin{bmatrix}
    \text{0}_{3\times3} \\
     \gamma \cdot \text{I}_{3\times3}
   \end{bmatrix}
\end{equation}

\noindent where $\Delta T$ is the sampling interval, $\text{I}_{3\times3}$ is the identity matrix of dimension $3 \times 3$ and $\text{0}_{3\times3}$ is the zero matrix of dimension $3 \times 3$. The parameters $\phi$ and $\gamma$ are further given by $\phi =  (1-\eta)$ and $\gamma = m^{-1} \Delta T$, and the parameter $\eta$ is used to model the air resistance. 

\subsection{Agent Sensing Model} \label{ssec:sensing_model}

\begin{figure*}
	\centering
	\includegraphics[width=\textwidth]{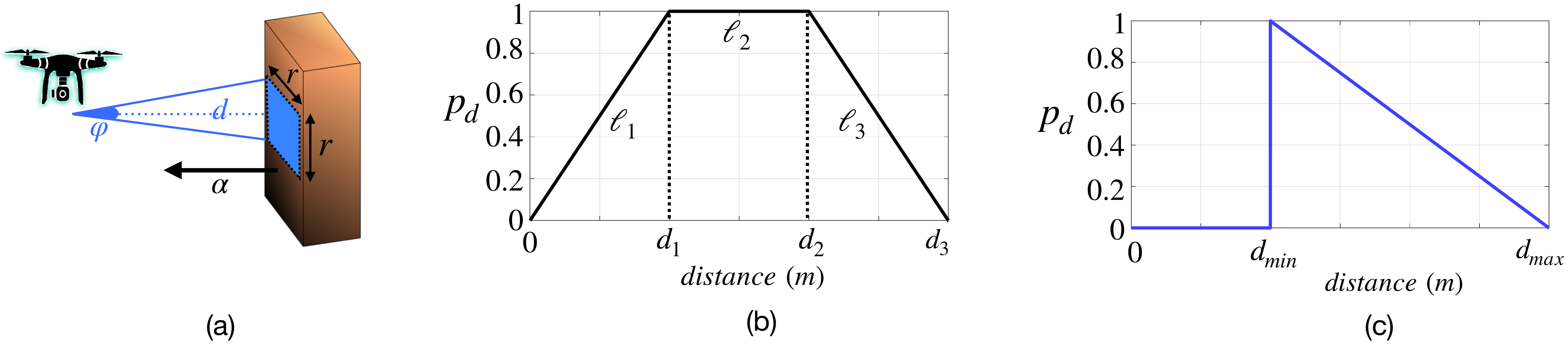}
	\caption{The figure illustrates the agent sensing model discussed in Sec. \ref{ssec:sensing_model}. a) the geometry of the UAV onboard camera FoV according to Eqn. (\ref{eq:camera_model}), b) an intuitive representation of the probability of detection $p_d(d)$ as a function of the distance between the agent and the face to be searched, c) the probability of detection model used in this work.}	
	\label{fig:sensing}
	\vspace{-0mm}
\end{figure*}

The agent is equipped with an onboard camera taking snapshots of the objects of interest in order to search for survivors or people in need. Without loss of generality, we assume in this work that the camera field of view  (FoV) angles in the horizontal and vertical axis are equal \cite{Petrides2017} and thus the projected FoV footprint on a planar surface is square with side length $r$ and given by:
\begin{equation}\label{eq:camera_model}
    r(d) = 2 d  \tan\left(\frac{\varphi}{2}\right)
\end{equation}
where $d$ denotes the distance in meters between the location of agent and the surface of the object that needs to be searched and $\varphi$ is the angle opening of the FoV according to the camera specifications. Thus the area of the FoV footprint at a distance $d$ is $r(d)^{2}$ meters. Before taking a snapshot of the object of interest the agent first aligns its camera with respect to the surface in such a way so that the optical axis of the camera (i.e., the viewing direction) is parallel to the normal vector ($\boldsymbol{\alpha}$) of the surface as depicted in Fig. \ref{fig:sensing}a. 

In order to search an object of interest the agent needs to take multiple snapshots (according to the size of the FoV as given by Eqn. \eqref{eq:camera_model}) such that each face of the object is completely included in the acquired images. The acquired images are then processed by an image processing module in order to determine the presence of people. The confidence of the \textit{search}-task (i.e., how well people are detected) mainly depends on the quality of the acquired frames and on the size of the FoV.

Intuitively, the FoV footprint increases as the distance between the agent's location and the surface to be searched increases, allowing the agent to capture a larger area of the object of interest. However, the amount of detail captured in those images inversely decreases with the size of the FoV and as a consequence the probability of detecting  people decreases due to insufficient pixel density. On the other hand, as the distance between the agent and the surface decreases the probability of detecting people in the captured frames increases, however the size of the FoV footprint decreases and thus less area is covered. That said, we define the confidence of the search task (i.e., the \textit{search confidence}) as: 

\begin{equation}\label{eq:conf}
q(d) \sim p_d(d) \times \frac{r^2(d)}{\text{m}^2}
\end{equation}

\noindent where $p_d(d) \in [0,1]$ is the probability of detecting people in images captured at distance $d$ meters from the object of interest and $r^2(d)$ denotes the area of the camera's FoV (in squared meters i.e., $\text{m}^2$) at distance $d$ from the object of interest as given in Eqn. (\ref{eq:camera_model}).% For the rest of the paper we will refer to $q(d)$ as the \textit{search confidence}.

We assume that the probability of detection is given by a piecewise linear function composed of 3 pieces $\ell_1$,  $\ell_2$ and  $\ell_3$ as shown in Fig. \ref{fig:sensing}b. Our assumptions here are the following a) when the agent is at distance $d=0$ from the object of interest, the probability of detecting people in the acquired frame should be zero. This is because the size of the camera FoV vanishes at $d=0$. Thereafter, we assume that as the distance between the agent and the object of interest increases the probability of detection increases until it reaches its maximum value at some distance $d_1$. In other words as the FoV size increases, the amount of information included in the acquired images increases as well, and as a result the probability of detection increases until it reaches it's maximum value, b) there is a range i.e., $d_1$ to $d_2$ in where the probability of detection remains constant and equal to the maximum value. In other words when the agent is at distance $d \in [d_1,d_2]$ the probability of detection does not depends on the FoV size (i.e., the pixel density in the acquired images is optimal with respect to the people detection process and further increase in the FoV size does not improve the results of this process) and c) after some distance $d_2$ the probability of detection drops as the FoV size increases. In this case the number of pixels that are used to represent people in the image frame drops significantly, and as a result people cannot be detected due to insufficient pixel density. Equivalently, the probability of detection $p_d(d)$ (based on the assumptions discussed above) can also be defined as: 
\begin{equation} \label{eq:pd}
p_d(d) = 
\begin{cases} 
   0 & , ~\text{if }~ d \le d_\text{min} \\
   \text{max} (0, ~ 1-\frac{d-d_\text{min}}{d_\text{max}-d_\text{min}}) & ,~ \text{if }~ d > d_\text{min}
\end{cases}
\end{equation}

\noindent where $d_\text{min}$ and $d_\text{max}$ are the minimum and maximum camera working distance for detecting people in the acquired frames. In order to see this first observe from Fig. \ref{fig:sensing}b that the agent has no incentive to position itself at a distance less than $d_1$ when maximizing the search confidence $q(d)$ since $q(d_1) > q(d), \forall d \in [0,d_1)$. For this reason the line piece $\ell_1$ disappears as shown in Fig. \ref{fig:sensing}c. Similarly, the line segment $l_2$ collapses to a point since the agent will always prefer $q(d_2)$ over $q(d), \forall d \in [d_1,d_2)$. Thus the final form of $p_d(d)$ which we use in this work is given by Eqn. (\ref{eq:pd}) and shown in Fig. \ref{fig:sensing}c.

In Sec. \ref{sec:search_planning} we show how we have incorporated the detection probability constraint into our mathematical programming formulation in order to generate optimal search plans that guide the UAV agent to search a specific object of interest while maintaining the required detection probability.

\subsection{Object Representation} \label{ssec:cuboids}
We consider a bounded 3D environment which contains two types of objects: a) obstacles that need to be avoided and b) objects of interest that need to be searched in 3D (i.e., searching the area of all their faces). All types of objects inside the surveillance environment are constructed using 3D primitives or building blocks. In this work these building blocks are rectangular cuboids of various sizes (referred to as simple cuboids). A rectangular cuboid is a convex hexahedron in three dimensional space which exhibits six rectangular faces (i.e., where each pair of adjacent faces meets in a right angle).
More specifically, let a plane $\mathcal{P}$ in 3D space be given by the set of points $x \in \mathbb{R}^{3}$ which satisfy:
\begin{equation}
    \mathcal{P} = \{x \in \mathbb{R}^{3} : \alpha^\top \cdot x = b\}
\end{equation}
where $\alpha^\top \cdot x$ is the dot product of the outward normal $\alpha^\top  = [\alpha_x,\alpha_y,\alpha_z]$ on the plane with the point $x$ and $b$ is a constant. The plane $\mathcal{P}$ divides $\mathbb{R}^3$ into two half-spaces i.e., the negative half-space $ \mathcal{P}^{-} = \{x \in \mathbb{R}^{3} : \alpha^\top \cdot x < b\}$ and the positive half-space $ \mathcal{P}^{+} = \{x \in \mathbb{R}^{3} : \alpha^\top \cdot x > b\}$. 

A rectangular cuboid $\mathcal{C}$ which is composed of six rectangular faces $f_i, i \in [1,..,6]$ can then be defined as the intersection of the six negative half-spaces:
\begin{equation} \label{eq:cuboid_def}
    \mathcal{C} = \left\{x \in \mathbb{R}^3 : x \in {\bigcap}_{i=1}^6 ~ \mathcal{P}^{-}_i\right\}
\end{equation}
where each pair of adjacent planes ($\mathcal{P}_i,\mathcal{P}_j, i \ne j$), which form the aforementioned half-spaces, intersect at right angles. The dimensions of the cuboid (i.e., length, height, depth)  are given by $\sigma(\mathcal{C}) = [\mathcal{C}_l,\mathcal{C}_h,\mathcal{C}_d]$.
The set of faces which is used to compose the cuboid $\mathcal{C}$ is denoted as $\mathcal{C}^f = \bigcup_{i=1}^{6} \{f_i\}$ and the dimensions (i.e., length and width) of each face is given by $\sigma(f_i) = [l,w] \subset \sigma(\mathcal{C})$.

Using simple cuboids we can construct more complex objects i.e., compound objects. A compound object $\tilde{\mathcal{C}}$ with length $|\tilde{\mathcal{C}}|$ is defined as the union of $|\tilde{\mathcal{C}}|$ simple cuboids:
\begin{equation}
    \tilde{\mathcal{C}} = {\bigcup}_{i=1}^{|\tilde{\mathcal{C}}|} ~ \mathcal{C}_i
\end{equation}
and the set of faces of $\tilde{\mathcal{C}}$ is given by $\tilde{\mathcal{C}}^f = \bigcup_{i=1}^{|\tilde{\mathcal{C}}|} \mathcal{C}_i^f$. That said, in this framework all objects (i.e., obstacles and objects of interests) are modeled either as simple cuboids or as compound objects. Our choice to use rectangular cuboids is twofold. First, it allows us to model a variety of object classes without compromising the level of detail, and secondly it allows us to formulate the search planning problem as a Mixed Integer Quadratic Program (MIQP) and solve it exactly using standard solvers. Additionally, we should point out that in this work we build on the assumption that a 3D map of the environment is readily available prior to planning i.e., the surveillance region has been 3D mapped \cite{Re,Rf,Rg,Rh} and subsequently the objects of interest and obstacles have been represented as cuboids.

\section{Unified Search Planning Framework} \label{sec:search_planning}

In this section we describe in detail the proposed \textit{Search Planning} framework which is used to automate the search phase of a SAR mission. More specifically, in this section we will discuss how the proposed framework takes into account the low-level mission constrains (i.e., UAV dynamical and sensing model discussed in Sec. \ref{sec:system_model}), the mission objectives and finally the mission specifications. At this phase all objects inside the environment are modeled as simple cuboids or compound objects. In addition, the human expert specifies a) the agent start position, b) the goal region $\mathcal{G}$ to be reached at the end of the mission, c) the planning horizon $T$ and finally d) the minimum required probability of detection $\mathcal{Q}$ to be maintained during the mission.

\subsection{Mission Objectives}\label{ssec:SARobj}
The main goal of a search mission is to search as efficiently as possible the target area for people in need. Thus the main objectives of the UAV-based search mission considered in this framework are a) the optimization of the mission's execution time (i.e., we would like to minimize the required search time while satisfying the mission constraints) and b) the optimization of the UAV's energy efficiency during the mission (i.e., to operate the UAV in an energy efficient manner). 

In order to handle the aforementioned objectives we define the required objective function as the weighted combination of two terms, namely: the \textit{Path Error Penalty} (PEP) and the \textit{Input Fluctuation Penalty} (IFP):
\begin{align}
    &\textit{Path Error Penalty:}\>\> e_T = \sum_{t=1}^T ||Hx_{t}-x_{\text{goal}}||^2_2 \label{eq:PEP} \\
    &\textit{Input Fluctuation Penalty:}\>\> \delta_T = \sum_{t=1}^{T-1} ||u_{t}-u_{t-1}||^2_2 \label{eq:IFP}
\end{align}

\noindent where $T$ is the planning horizon (i.e., the total amount of time allocated for the mission), $H$ is a matrix which extracts the position vector from the agent's state vector and $x_\text{goal} \in \mathcal{G}$ is the location of a point which belongs to the goal region and which the agent must reach at the end of the mission.

As we can observe from Eqn. (\ref{eq:PEP}), the \textit{Path Error Penalty} minimizes the sum of errors between the position of the agent $Hx_t$ and the desired goal location $x_\text{goal}$, which effectively drives the agent to finish the mission inside the goal region. Additionally, this objective forces the agent to reach the goal position as soon as possible i.e., at the earliest time-step. For this reason, the minimization of PEP also minimizes the search operation execution time. 

On the other hand, the \textit{Input Fluctuation Penalty} in Eqn. (\ref{eq:IFP}) is used in order to minimize the fluctuations between consecutive control inputs thus leading to smoother trajectories (i.e., smooth trajectories with less abrupt changes). Additionally, in this work we consider that the minimization of the UAV's control input variation is directly related to the reduction of the UAV's energy consumption. We assume that by reducing the variation between consecutive controls, abrupt changes in the control input are eliminated, leading to more energy efficient operation. Overall, the objective function of the search mission is defined as:
\begin{equation}\label{eq:SAR_objective}
    \underset{u_{0:T-1}}{\min} ~ h(x_{1:T},u_{0:T-1})= w_1 e_T + w_2 \delta_T
\end{equation}

\noindent where the weights $w_1$ and $w_2$ are chosen by the human operator according to the mission requirements and in essence determine the emphasis given between the mission execution time and the mission energy efficiency. 

\begin{figure*}
	\centering
	\includegraphics[width=\textwidth]{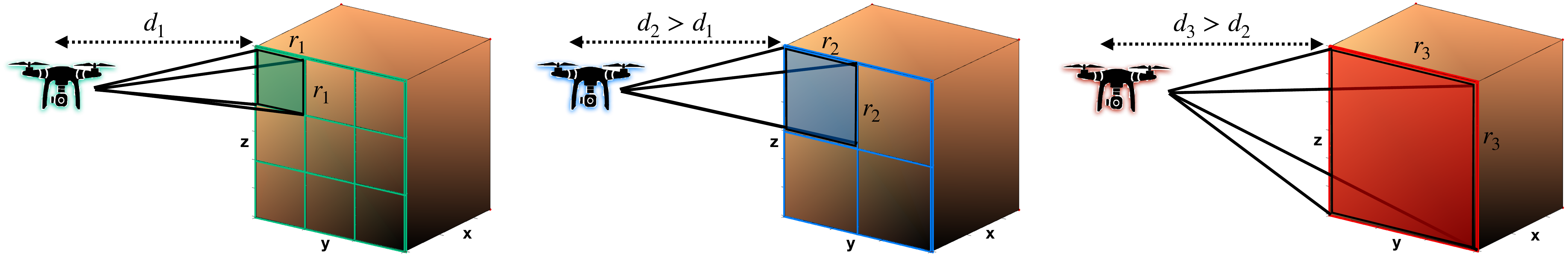}
	\caption{The figure illustrates the projected FoV shape (i.e., $r_i \times r_i$ square) as a function of the distance (i.e., $d_i$) between the UAV agent and the object of interest. Assuming the distance $d_i$ is kept constant, the UAV agent must scan each cell of the shown grid in order to search the total area of the face.}	
	\label{fig:fov_size}
	\vspace{-0mm}
\end{figure*}

\subsection{3D Search Task} \label{ssec:3d_search_task}
We can now describe how the proposed framework handles the 3D search task. More specifically, the objective here is to devise a search plan (i.e., a UAV trajectory) that by taking into account the UAV sensing model discussed in Sec. \ref{ssec:sensing_model}, will guide the UAV agent to search all faces (i.e., use the onboard camera module to scan the total area of each face) of an object of interest (which is represented as a simple cuboid $\mathcal{C}$ or as a compound object $\tilde{\mathcal{C}}$). To further illustrate this idea observe in Fig. \ref{fig:fov_size} that the camera FoV size i.e., a square $r \times r$, depends on the distance $d$ between the UAV agent and the object of interest. Assuming that the UAV maintains its distance $d$ from the object of interest, a particular face is searched when the UAV visits all cells of the grid shown in Fig. \ref{fig:fov_size}. To implement the 3D search task, we compute the aforementioned grid for multiple values of the distance $d$ and we generate 3D zones which are composed of simple cuboids. We then use mathematical programming techniques to guide the UAV agent through the generated 3D zones. By doing so, we constrain the position of the UAV over time to reside inside a particular set of cuboids which results in FoV projections which include the area of each cell, thus accomplishing the searching of the whole face of the object of interest. This process is performed from each face that needs to be searched. 

In essence, we discretize the area around the object of interest we wish to search by creating 3D zones at various distances according to the agent sensing model. Then we use mathematical programming techniques to guide the UAV through these 3D zones in order to search the object of interest. Next we discuss the details of the proposed 3D searching technique.

\subsubsection{3D Zone Construction} \label{ssec:3dzone}
A 3D zone $\tilde{\mathcal{Z}}$ with length $|\tilde{\mathcal{Z}}|$ is a compound object (i.e., a union of simple cuboids $\mathcal{Z}$) defined as:
\begin{equation}
    \tilde{\mathcal{Z}} = {\bigcup}_{i=1}^{|\tilde{\mathcal{Z}}|} ~ \mathcal{Z}_i
\end{equation}

\noindent which is created by the proposed framework around each object of interest to aid the 3D search task. All the simple cuboids that belong to $\tilde{\mathcal{Z}}$ are of the same size i.e., $\sigma(\mathcal{Z}_1) = \sigma(\mathcal{Z}_2) = \cdots = \sigma(\mathcal{Z}_{|\tilde{\mathcal{Z}}|})$. Additionally, these simple cuboids  $\mathcal{Z}_i$ have size of the form $\sigma(\mathcal{Z}_i) = [r,r,D]$ i.e., at least one pair of opposite faces is square. Let us denote one of the faces $f_\mathcal{Z} \in \mathcal{Z}^f$ (of the cuboid $\mathcal{Z}$) with size $r \times r$ as $\hat{f}^r_\mathcal{Z}$ (we have dropped the indexing on the cuboids since all cuboids in a zone are the same).

The value of $r$ of $\hat{f}^r_\mathcal{Z}$ is determined by Eqn. (\ref{eq:camera_model}) i.e., $r$ is given by the side length of the camera FoV when the agent is at distance $d$ from an object of interest. On the other hand, the value of $D$ determines the depth of the cuboid. The parameter $D$ also determines the depth of the zone $\tilde{\mathcal{Z}}$ thus for this reason it is also referred to as the zone depth $\tilde{\mathcal{Z}}_D$.

The length $|\tilde{\mathcal{Z}}|$ of a zone $\tilde{\mathcal{Z}}$ (i.e., how many cuboids the zone contains) is determined by the number of cuboids that need to be placed around the object of interest to enable the UAV agent to search all its faces. In essence the number of cuboids contained in a zone is equal to the number of grid cells (i.e., as shown in Fig. \ref{fig:fov_size}) times the number of faces that need to be searched.

Let the object of interest to be searched be denoted by the simple cuboid $\mathcal{R}$ with size $\sigma(\mathcal{R}) = [\mathcal{R}_l, \mathcal{R}_w, \mathcal{R}_d] $ and set of faces given by $\mathcal{R}^f$. This cuboid contains 3 pair of faces with sizes $\mathcal{R}_l \times \mathcal{R}_w$,  $\mathcal{R}_l \times \mathcal{R}_d$ and $\mathcal{R}_w \times \mathcal{R}_d$ respectively. To enable the 3D search of $\mathcal{R}$ from a particular distance $d$ we generate a zone of cuboids around $\mathcal{R}$ and we guide the UAV agent through the zone. More specifically, for each face $f_i \in \mathcal{R}^f$, with size $\sigma(f_i) \in \{[\mathcal{R}_l,\mathcal{R}_w], [\mathcal{R}_l, \mathcal{R}_d], [\mathcal{R}_w, \mathcal{R}_d]\}$ we find how many non-overlapping faces $\hat{f}^r_\mathcal{Z} \in \mathcal{Z}^f$ with square shape $r(d) \times r(d)$ of the cuboid $\mathcal{Z} \in \tilde{\mathcal{Z}}$ with size $\sigma(\mathcal{Z}) = [r(d),r(d),D]$ are needed to fully cover $f_i$. We should point out here that the above procedure assumes that $\hat{f}^r_\mathcal{Z} \parallel f_i$ i.e., a cuboid $\mathcal{Z}$ is placed in such a way so that its face $\hat{f}^r_\mathcal{Z}$ is parallel to the face $f_i$ that must be searched. The above procedure is repeated for different values of $d$. 

What we have discussed so far allows us to determine the length of a zone $\tilde{\mathcal{Z}}$. Next we discuss how we determine the sizes of the cuboids inside the zones and how we choose the number of zones. 
\begin{figure*}
	\centering
	\includegraphics[width=\textwidth]{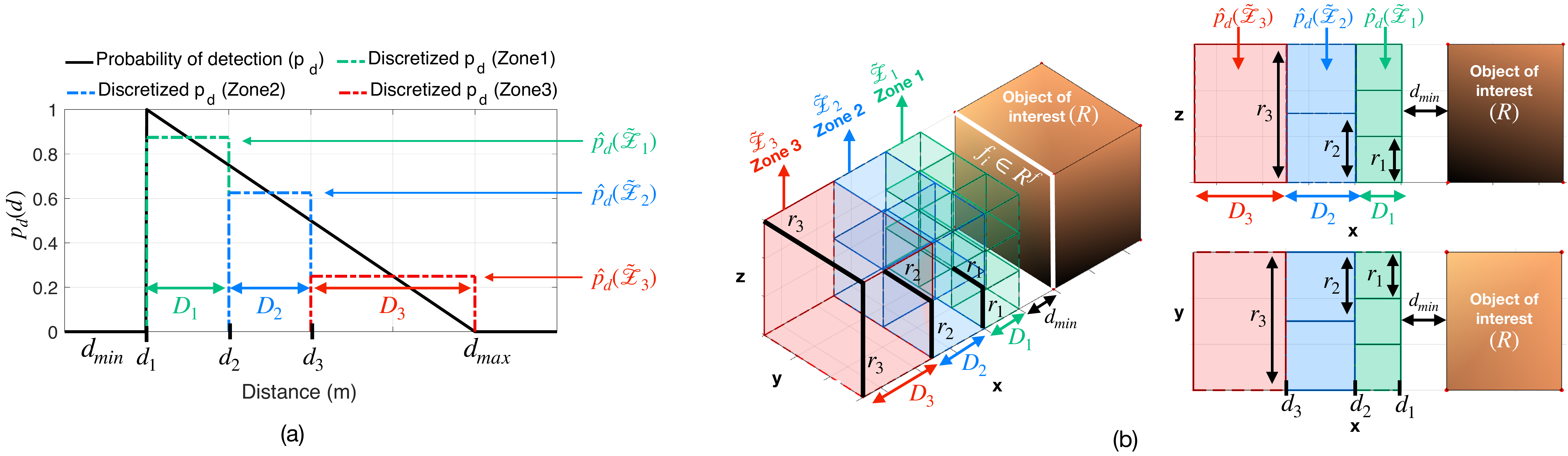}
	\caption{The figure illustrates the 3D zone construction step discussed in Sec. \ref{ssec:3d_search_task} which enables the 3D search planning of an object of interest. a) The figure shows the discretization of the detection probability, i.e., Eqn. (\ref{eq:pd}), into $\mathcal{N}=3$ zones $\tilde{\mathcal{Z}}_1,~ \tilde{\mathcal{Z}}_2$ and $\tilde{\mathcal{Z}}_3$ depicted in green, blue and red color respectively. The 3 zones have depth values $D_1, D_2$ and $D_3$ as illustrated and detection probability values $\tilde{p_d}(\tilde{\mathcal{Z}}_1),~ \tilde{p_d}(\tilde{\mathcal{Z}}_2)$ and $\tilde{p_d}(\tilde{\mathcal{Z}}_3)$. Moreover, the 3 zones $\tilde{\mathcal{Z}}_1,~ \tilde{\mathcal{Z}}_2$ and $\tilde{\mathcal{Z}}_3$ are placed at distances of $d_1, d_2$ and $d_3$ respectively from the object of interest. b) The figure shows the zone construction step (for the face $f_i \in \mathcal{R}^f$) of the object of interest $R$. Zones 1, 2 and 3 have lengths of $|\tilde{\mathcal{Z}}_1| = 9$,~ $|\tilde{\mathcal{Z}}_2| = 4$ and $|\tilde{\mathcal{Z}}_3| = 1$ as illustrated in this example. The size of the cuboids in each zone i.e., $\sigma(\mathcal{Z}_i)=[r_i,r_i,D_i]$ is determined by the size of the FoV footprint at distance $d_i$ (computed using Eqn. (\ref{eq:camera_model})) from the object of interest $\mathcal{R}$ and from the depth of the zone. Please note that in the illustration above we show the zone construction step for only one face. In order to search the object in 3D the 3 zones will be expanded to cover all 5 faces. }	
	\label{fig:zones}
\end{figure*}

As we have briefly discussed we create 3D zones around the object of interest in order to allow the UAV agent to pass through them and search in 3D all the faces of the object. In essence these 3D zones can be considered as a discretized version of the agent search confidence i.e., Eqn. (\ref{eq:conf}). This discretization is employed in this work because it allows us to easily incorporate the functionality supported by Eqn. (\ref{eq:camera_model}) and Eqn. (\ref{eq:pd}) into our mathematical programming framework and use  standard solvers to tackle the derived mixed integer quadratic program (MIQP) program. That said, the number of zones $\mathcal{N}$ to be created is user-defined and in essence determines the granularity of the discretization of the space around the object of interest. 

To be more precise, we assign to each zone $\tilde{\mathcal{Z}}_i$ a depth value $D_i$ and a value $\hat{p_d}(\tilde{\mathcal{Z}}_i)$ for the probability of detection by quantizing the domain and range of Eqn. (\ref{eq:pd}) into $\mathcal{N}$ components. In essence, the domain of Eqn. (\ref{eq:pd}) is partitioned into $\mathcal{N}$ regions and the length of each of those regions determines the depth $D_i$ of each zone. Equivalently, the values of $p_d(d)$ at the partitioned regions are assigned as the values of the detection probability achievable in each zone i.e., $\hat{p_d}(\tilde{Z}_i)$ as illustrated in Fig. \ref{fig:zones}a.
That said, the cuboids in each zone $i$ have size $[r_i, r_i, D_i]$ where $D_i$ is determined by the zone depth and $r_i$ is derived by computing the distance between the cuboid and the object of interest and applying Eqn. (\ref{eq:camera_model}). 

For the 3D search task we generate a set of 3D zones $\tilde{\mathcal{Z}}_i$, around the object of interest, each of which is placed at a distance $d_i$ from the object. Each zone is assigned a detection probability value which corresponds to the detection probability that is achievable at distance $d_i$ according to Eqn. (\ref{eq:pd}). Additionally, each zone has depth $D_i$ which has been computed by partitioning the domain of Eqn. (\ref{eq:pd}) into $\mathcal{N}$ regions. Finally, each zone $\tilde{\mathcal{Z}}_i$ is composed of $|\tilde{\mathcal{Z}}_i|$ cuboids of the same size. The length of each zone $|\tilde{\mathcal{Z}}_i|$ is determined by how many simple cuboids are needed to cover the object faces. The size of the cuboids is determined by the size of the FoV footprint at distance $d_i$ from the object and from the depth of the zone. The procedure discussed above is illustrated in Fig. \ref{fig:zones}b.

Finally, the agent can search in 3D the object of interest by passing through the appropriate 3D zone (i.e., visit all cuboids contained in the zone) according to the required detection probability which is specified by the human operator. The way that the detection probability is being discretized and the formation of the various zones around the objects of interest is according to the mission specifications and the mission goals i.e., the number and size of the 3D zones are such that the mission-specific required detection probability levels are captured.

\begin{figure}
	\centering
	\includegraphics[scale=0.25]{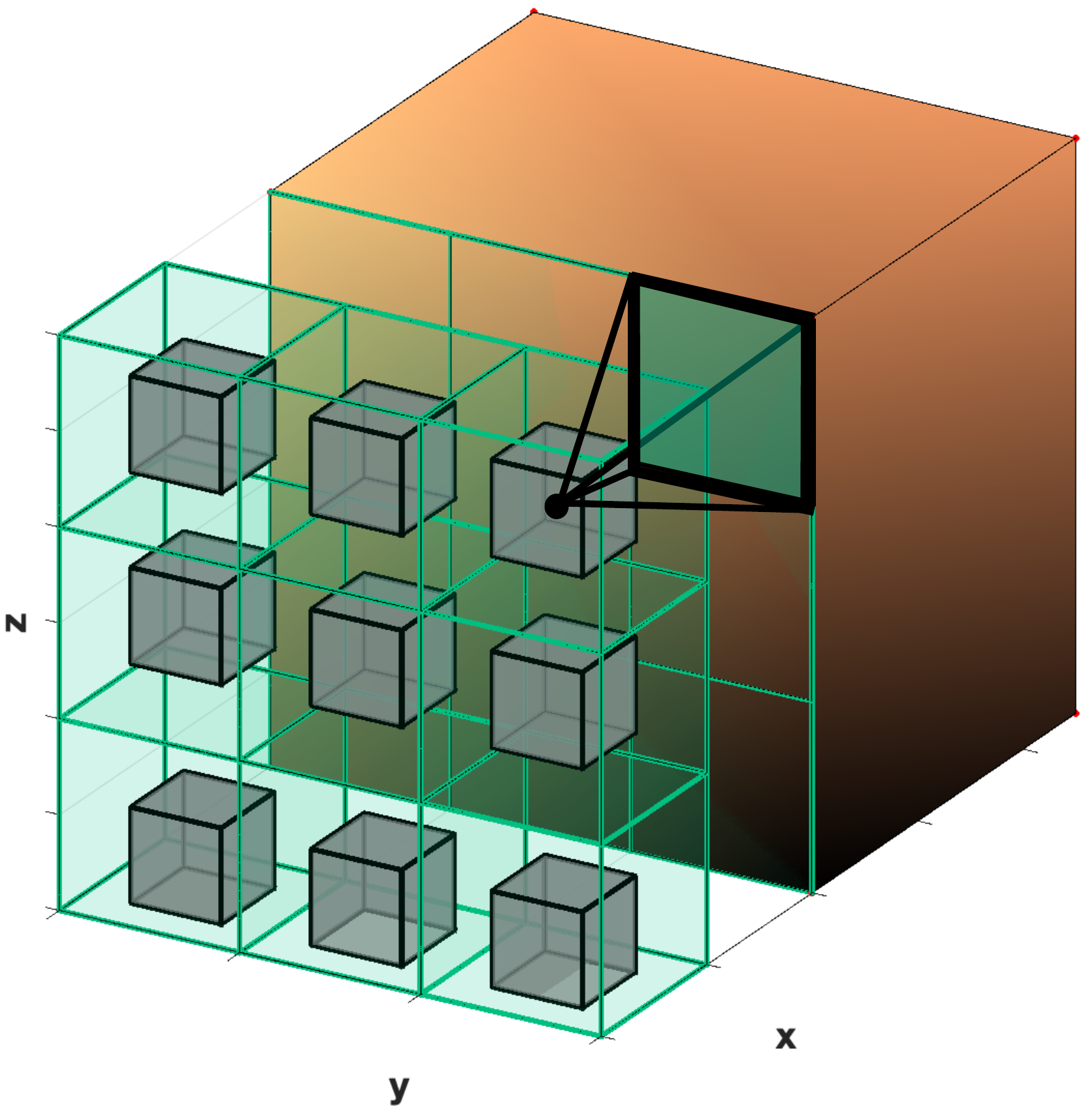}
	\caption{The figure illustrates the interior cubes (gray) centered inside the cuboids (green). By passing through each cube the UAV agent makes sure that the whole area of the face is searched i.e., the projected FoV from within a cube encloses the area of the corresponding grid cell on the face's surface as shown in the figure.}	
	\label{fig:waypoints}
	\vspace{-0mm}
\end{figure}

We should point out here that in order to make sure that the projected FoV of the UAV agent (at a particular time instance when the agent resides inside a cuboid) completely covers the corresponding grid cell on the surface of the object of interest (e.g., Fig. \ref{fig:fov_size}), in our implementation we require the agent to pass approximately from the center of each cuboid. Equivalently, we require the agent to pass through an interior cube centered inside each cuboid. This is depicted in Fig. \ref{fig:waypoints}. In Sec. \ref{ssec:3D_con} we show how we encode the above constraints. 

%For the rest of the paper we will assume that whenever the agent passes through some cuboid $\mathcal{Z} \in \tilde{\mathcal{Z}}$ it also passes through its interior cube.
To summarize, we generate 3D grids at various distances with respect to the faces that need to be searched. By guiding the agent to visit each cell of this 3D grid, we make sure that the total projected FoV captures the whole area of the face. Which 3D grid the agent will visit depends on the required detection probability. 

\subsubsection{3D Search Constraints}\label{ssec:3D_con}
We can now describe how we have encoded the 3D search task into mathematical programming constraints (shown in Program \ref{alg:3d_search_task}). In essence, we would like to make sure that the UAV agent will pass through the appropriate zone $\tilde{\mathcal{Z}}_i$ during the mission (i.e., traversing each interior cube of every simple cuboid $\mathcal{Z} \in \tilde{\mathcal{Z}}_i$ inside the appropriate zone) thus searching in 3D the object of interest. We should point out here that from which zone $\tilde{\mathcal{Z}}_i$ the agent will go through depends on the user defined detection probability $\mathcal{Q}$.
In order to simplify the notation, and without loss of generality, we describe the 3D search task for one object of interest. The constraints in Eqn. (\ref{eq:S_1})-(\ref{eq:S_6}) which implement the 3D search task however, can be easily extended to handle multiple objects of interest. 

% -------------------------------- 3D Search Constraints ----------------------
\begin{algorithm}
\caption{3D Search Task}
\label{alg:3d_search_task}
\begin{align}
& \boldsymbol{\alpha}^\top_{lci} Hx_{t} + (M-b_{lci}) z_{tlci} \le M, ~\forall t,l,c,i \label{eq:S_1}\\
& L \tilde{z}_{tci} - \sum_{l=1}^{L} z_{tlci} \le 0, ~ \forall t,c,i \label{eq:S_2}\\ 
& -\sum_{t=1}^T \tilde{z}_{tci} + \hat{z}_i \le 0, ~ \forall i, c \label{eq:S_3}\\
& -\sum_{t=1}^T \sum_{c=1}^{|\tilde{\mathcal{Z}}_i|} \tilde{z}_{tci} + |\tilde{\mathcal{Z}}_i| \hat{z}_i \le 0 , ~ \forall i \label{eq:S_4}\\
& -\sum_{i=1}^\mathcal{N} \hat{z}_i \hat{p_d}(\tilde{\mathcal{Z}}_i)  \le -\mathcal{Q} , ~ \forall i \label{eq:S_5}\\
&  \sum_{i=1}^\mathcal{N} \hat{z}_i \le 1 , ~ \forall i \label{eq:S_6}
\end{align}
\end{algorithm}
%% ---------------------------------------------------------------------------

Let us assume that the 3D zone construction procedure discussed earlier has generated zones $\tilde{\mathcal{Z}}_i, i \in [1,..,\mathcal{N}]$ (around the object of interest) where each zone $\tilde{\mathcal{Z}}_i$ is composed of cuboids $\mathcal{Z}_c, c \in [1,..,|\tilde{\mathcal{Z}}_i|]$ where each cuboid $\mathcal{Z}_c$ contains an interior cube $\mathcal{Y}_c$ which has $L=6$ faces.

 Assuming that the total duration of the mission can take up to $T$ time steps, we define the binary variable $z_{tlci}$ which at time $t \in [1,..,T]$ points to the face $l \in [1,..,L]$ of the interior cube $\mathcal{Y}_c, c \in [1,..,|\tilde{\mathcal{Z}}_i|]$ of cuboid $\mathcal{Z}_c$ of zone $i \in [1,..,\mathcal{N}]$. The binary variable $\tilde{z}_{tci}$ points to the interior cube $\mathcal{Y}_c, c \in [1,..,|\tilde{\mathcal{Z}}_i|]$ of cuboid $\mathcal{Z}_c$ of zone $i$ at time $t$ and finally the binary variable $\hat{z}_i$ activates the $i_\text{th}$ zone.

With the above definitions and in order to accomplish the 3D search task, we first check with constraint Eqn. (\ref{eq:S_1}) whether the agent position $Hx_t$ resides inside the negative half-space $\mathcal{P}^{-}_{lci} = \{Hx_t \in \mathbb{R}^{3} : \boldsymbol{\alpha}^\top_{lci} \cdot Hx_t \le  b_{lci}\}$ created by the plane $\mathcal{P}_{lci}$ of the $l_\text{th}$ face of the interior cube $\mathcal{Y}_c, c \in [1,..,|\tilde{\mathcal{Z}}_i|]$ of cuboid $\mathcal{Z}_c$ of zone $i$. If the constraint is satisfied the binary variable $z_{tlci}$ is activated, otherwise we make sure that the inequality is still valid using big-M techniques (i.e., $M$ is a large positive constant to make sure that the inequality is valid at all times). Then constraint in Eqn. (\ref{eq:S_2}) checks whether the agent resides inside the interior cube $\mathcal{Y}_c$ of cuboid $\mathcal{Z}_c$ of zone $i$ at time-step $t$. In essence, this constraint implements Eqn. (\ref{eq:cuboid_def}) and activates the binary variable $\tilde{z}_{tci}$ if the agent position resides inside the interior cube of the desired cuboid.

The constraint in Eqn. (\ref{eq:S_3}) makes sure that each of the interior cubes of all cuboids, which belong to a specific zone, will be visited at least once during the duration $T$ of the mission and subsequently the constraint in Eqn. (\ref{eq:S_4}) ensures that the agent will visit all $|\tilde{\mathcal{Z}}_i|$ cuboids and their corresponding interior cubes of the selected zone $\tilde{\mathcal{Z}}_i$ indicated by the binary variable $\hat{z}_i$. These two  constraints in essence make sure that the object will be searched completely in 3D (i.e., by visiting all different cuboids in a zone, all faces of the objects are searched). 

As we have already explained previously in this section each zone is assigned a detection probability value $\hat{p_d}(\tilde{\mathcal{Z}}_i)$ (where in this notation $\hat{p_d}(\tilde{\mathcal{Z}}_i)$ denotes the detection probability value indicated by the binary variable $\hat{z}_i$) which depends on the distance between the zone and the object of interest. The constraint in Eqn. (\ref{eq:S_5}) selects the zone $i$ which achieves the specified detection probability $\mathcal{Q}$ requested by the human operator and activates the binary variable $\hat{z}_i$. Finally, the constraint in Eqn. (\ref{eq:S_6}) makes sure that only one zone is selected. To summarize, Program \ref{alg:3d_search_task} makes sure that the appropriate 3D zone is selected which achieves a detection probability at least $\mathcal{Q}$ and that all the cuboids inside the selected zone are visited. 

\subsection{Obstacle Avoidance Task} \label{ssec:obstacle_avoidance}
The obstacles considered in this work are modeled as simple cuboids $\mathcal{C}$ or compound objects $\tilde{\mathcal{C}}$. The objective of the obstacle avoidance task is to constrain the generated path within the free-space region, thus enabling the agent to avoid collisions with the obstacles. Let us assume for the sake of notational convenience that the obstacles are modeled as simple cuboids $\mathcal{C}_\psi, \psi \in \Psi$ where $\Psi \subset \Xi \cup J$ i.e., the agent should avoid collisions not only with the obstacles $\Xi$ in the environment but also with the objects of interest $J$. Thus, the obstacle avoidance task is accomplished when the agent position $Hx_t$ satisfies:
\begin{equation}
    Hx_t \notin \mathcal{C}_\psi,~\forall \psi \in \Psi, t \in [1,..,T]
\end{equation}

Thus we need to ensure that the agent resides outside the cuboid(s) that form the obstacles at all times. The obstacle avoidance constrains can thus be written with the constraints in Eqn. (\ref{eq:O_1})-(\ref{eq:O_2}):

% -------------------------------- Obstacle Avoidance ---------------------------------------
%\begin{algorithm}
%\caption{MIQP for Object Monitoring}
%\label{alg:obstacle_avoidance}
\begin{align}
% ---- Obstacles -----
& -\boldsymbol{\alpha}^\top_{\psi l} Hx_{t} - M\varepsilon_{t \psi l} \le -b_{\psi l},~\forall t,\psi,l \label{eq:O_1}\\
& \sum_{l=1}^{L} \varepsilon_{t\psi l} \le  L-1, ~ \forall t,\psi \label{eq:O_2}
\end{align}
%\end{algorithm}
%% ---------------------------------------------------------------------------

\noindent where we use the binary variable $\varepsilon_{t\psi l}$ to index the face $l \in [1,..,L]$ of obstacle $\psi \in \Psi$ at time-step $t \in [1,..,T]$. The constraint in Eqn. (\ref{eq:O_1}) assigns a zero value to $\varepsilon_{t\psi l}$ when the agent position $H(x_t)$ belongs to the positive half-space formed by the plane $\mathcal{P}_{\psi l} = \{Hx_t \in \mathbb{R}^{3} : \boldsymbol{\alpha}^\top_{\psi l} \cdot Hx_t =  b_{\psi l}\}$. Otherwise, the agent position belongs to the negative half-space thus $\varepsilon_{t\psi l}$ is activated to make sure that the inequality is satisfied when $M$ is a large constant. Then the constraint in Eqn. (\ref{eq:O_2}) makes sure that a collision is avoided for all obstacles and time-steps by ensuring that the number of times $\varepsilon_{t\psi l}$ is activated for a specific $\psi$ and $t$ is less than $L-1$. In other words a collision is avoided when $\exists~ l \in [1,..,L]: \boldsymbol{\alpha}^\top_{\psi l} Hx_{t} > b_{\psi l}$.

\subsection{Reach Goal Region Task} \label{ssec:reach_goal}
The objective of this task is to allow the agent to reach the goal region (e.g., landing pad or re-charging station) during a pre-specified time-window $[\tau,..,T]$ or at the end of the mission $T$. Let the goal region be represented by a user specified simple cuboid $\mathcal{G}$ with $L=6$ faces, then the objective is to make sure that the UAV agent will reside inside $\mathcal{G}$ at the end of the mission or reach $\mathcal{G}$ during the time-window $[\tau,..,T]$. In essence, the constrains that are used to mathematically implement this task are exactly the opposite of those in obstacle avoidance which have been described in the previous paragraph. More specifically, the reach goal task is given by the constraints in Eqn. (\ref{eq:G_1})-(\ref{eq:G_3}):

% -------------------------------- P2 ---------------------------------------
%\begin{algorithm}
\begin{align}
% ---- Goal ------
& \boldsymbol{\alpha}^\top_{l} Hx_{t} + (M-b_{l})y_{tl} \le M, ~\forall t,l \label{eq:G_1} \\
& L\tilde{y}_{t} - \sum_{l=1}^{L} y_{tl} \le 0, ~ \forall t \label{eq:G_2}\\
& -\sum_{t=\tau}^T \tilde{y}_t \le -1 \label{eq:G_3}
\end{align}
%\end{algorithm}
%% ---------------------------------------------------------------------------

More specifically, the binary variable $y_{tl}, ~ t \in [1,..T], l \in [1..,L]$ which appears in Eqn. (\ref{eq:G_1}) is activated when the agent position resides inside the negative half-space formed by the $l_\text{th}$ plane with equation $\boldsymbol{\alpha}^\top_{l} Hx_{t} \le b_{l}$. Then the binary variable $\tilde{y}_{t}$ in Eqn (\ref{eq:G_2}) is activated when the agent is inside the goal region at time-step $t$. Finally, the constraint in Eqn. (\ref{eq:G_3}) ensures that the goal region is visited at least once inside the time-window $[\tau,..,T]$. 

The final MIQP for the entire 3D search planning phase is shown in Program \ref{alg:P2}, where the mission objective is due to Eqn. (\ref{eq:SAR_objective}), the agent dynamics are according to Eqn. (\ref{eq:agent_dynamics}) as described in Sec. \ref{sec:system_model} assuming a known initial state $x_0$ and a fixed planning horizon $T$. Then the constraints in Eqn. (\ref{eq:S_1})-(\ref{eq:S_6}) are responsible for the 3D search task of an object of interest, Eqn. (\ref{eq:O_1})-(\ref{eq:O_2}) implement the obstacle avoidance task and finally Eqn. (\ref{eq:G_1})-(\ref{eq:G_3}) guide the agent to the goal region.

% -------------------------------- P2 ---------------------------------------
\begin{algorithm}
\caption{Unified Search Planning Framework}
\label{alg:P2}
\begin{align}
% ---- Obj ------
\mathrm{(P2)}\>\>\>  &\underset{u_{0:T-1}}{\min} ~ h(x_{1:T},u_{0:T-1})= w_1 e_T + w_2 \delta_T \label{eq:objective_P2} \notag \\
% ---- Dynamics ------
\mathrm{s.t.} \>\>\>\> &  x_t = \Phi^t x_0 + \sum_{\tau=0}^{t-1} \Phi^\tau \Gamma [u_{t-\tau-1}-u_g], ~\forall t \in [1,..,T]  \notag \\
& \text{Eqn.} ~ (\ref{eq:S_1})-(\ref{eq:S_6})\>\>\>\>(\textit{3D search task}) \notag\\
& \text{Eqn.} ~ (\ref{eq:O_1})-(\ref{eq:O_2})\>\>\>\>(\textit{Obstacle avoidance task}) \notag\\
& \text{Eqn.} ~ (\ref{eq:G_1})-(\ref{eq:G_3})\>\>\>\>(\textit{Reach goal region task}) \notag
\end{align}
\end{algorithm}
\vspace{-0mm}
%% ---------------------------------------------------------------------------

In this work we have presented a search planning framework for searching objects of interest in 3D. We generate plans that guide an autonomous UAV agent to pass through a series of artificially generated cuboids, in order to search the faces of an object of interest with the specified detection probability, while at the same time avoiding collisions with obstacles in the environment. The proposed search-planning framework is flexible and it can be generalized for a variety of planning scenarios. An overview of the main requirements and key features of the proposed approach are the following: (a) The proposed method requires a known (i.e., size and location) cuboid representation of the objects of interest and obstacles in the environment. This cuboid representation however, can be easily be obtained from environmental 3D maps i.e., \cite{Re,Rf,Rg,Rh}. (b) The 3D zone construction step discussed in Sec. \ref{ssec:3dzone} and illustrated in Fig. \ref{fig:zones} takes place prior to mission planning and requires a known set of detection probability values that will be used during search planning. This set of detection probability values is mission specific and determined by the mission control according to the SAR mission requirements (i.e., depending on the severity or the importance of the search task). (c) As it will be shown in Sec. \ref{sec:Evaluation}, the proposed framework can then be used to generate the optimal trajectory which guides the UAV agent through the generated 3D zones in such a way so that the object of interest is searched according to the user specified detection probability. In addition as it will be demonstrated in the evaluation, the mission objective can be tuned accordingly (i.e., with respect to the SAR mission requirements) to configure the emphasis given between the mission execution time and the UAV energy efficiency. 

Although, in this work we have focused in the task of searching the faces of an object of interest, the proposed framework can be easily extended to support other tasks which can potentially arise in a SAR mission. These include searching for survivors inside buildings and under rubble and debris of demolished buildings in the event of earthquakes or other man-made or natural disasters.

In the case of searching for survivors inside a building, the 3D zone generation step described in Sec. \ref{ssec:3dzone}, and the 3D search constraints i.e., Sec. \ref{ssec:3D_con} can be applied indoors, by generating and placing artificial cuboids in the locations/areas that need to be searched by the UAV agent while optimizing the objective function of Eqn. \eqref{eq:SAR_objective} i.e., mission execution time and UAV energy efficiency. Subsequently, the obstacle avoidance constrains of Sec. \ref{ssec:obstacle_avoidance} can be applied to account for the walls of the building and the reach goal constraints i.e., Sec. \ref{ssec:reach_goal} can be used in order for the UAV to exit the building at the end of the mission. Moreover, in the case where the UAV agent is equipped with thermal/infra-red cameras and advanced imaging sensors, the proposed approach can be utilized to search for survivors under rubble and debris, by guiding the UAV agent through a series of artificially generated cuboids placed in selected locations above and around the rubble, acting as waypoints, allowing the UAV agent to detect heat signatures at a distance from above.

This concludes the description of the proposed search planning framework. Next we present the experimental evaluation of the proposed approach using a variety of simulated search scenarios.

\section{Evaluation} \label{sec:Evaluation}

\subsection{Experimental Setup}
To evaluate the performance of the proposed search planning framework we conduct several synthetic experiments. In each experiment we evaluate the proposed approach either qualitatively or quantitatively and we discuss its strengths and weaknesses. The experimental evaluation is divided into four parts. In the first part we present the overall behavior of the proposed search planning framework focusing on the 3D search task (i.e. Sec. \ref{ssec:3d_search_task}) including the 3D Zone Construction and 3D search constraints. Next, in part 2 we show how the mission objective discussed in Sec. \ref{ssec:SARobj} affects the search planning behavior of the agent. In part 3 of our evaluation, we demonstrate the performance of proposed search planning framework in the presence of obstacles and finally we conclude with a discussion regarding the computational complexity of the proposed framework and future directions.

The experimental setup used for the evaluation of the proposed system is as follows: The agent dynamics are expressed by Eqn. (\ref{eq:agent_dynamics}) with $\Delta T = 1$s, agent mass $m=3.35$kg and air resistance coefficient $\eta = 0.2$. The applied control input (i.e., input force) $u_t = [\text{u}_x, \text{u}_y, \text{u}_z] $ is bounded within the intervals $[-35, 35]$N, $[-35, 35]$N and $[-10, 35]$N in $x$, $y$ and $z$ dimension respectively, the agent velocity  $\dot{\text{x}} = [\nu_x,\nu_y,\nu_z]$ is bounded in each dimension within the interval $[-15, 15]$m/s and the agent FoV angle $\phi$ is 60deg. Simulations were conducted on an 2GHz dual core CPU running the Gurobi V8 MIQP solver. 

\begin{figure}
	\centering
	\includegraphics[scale=0.3]{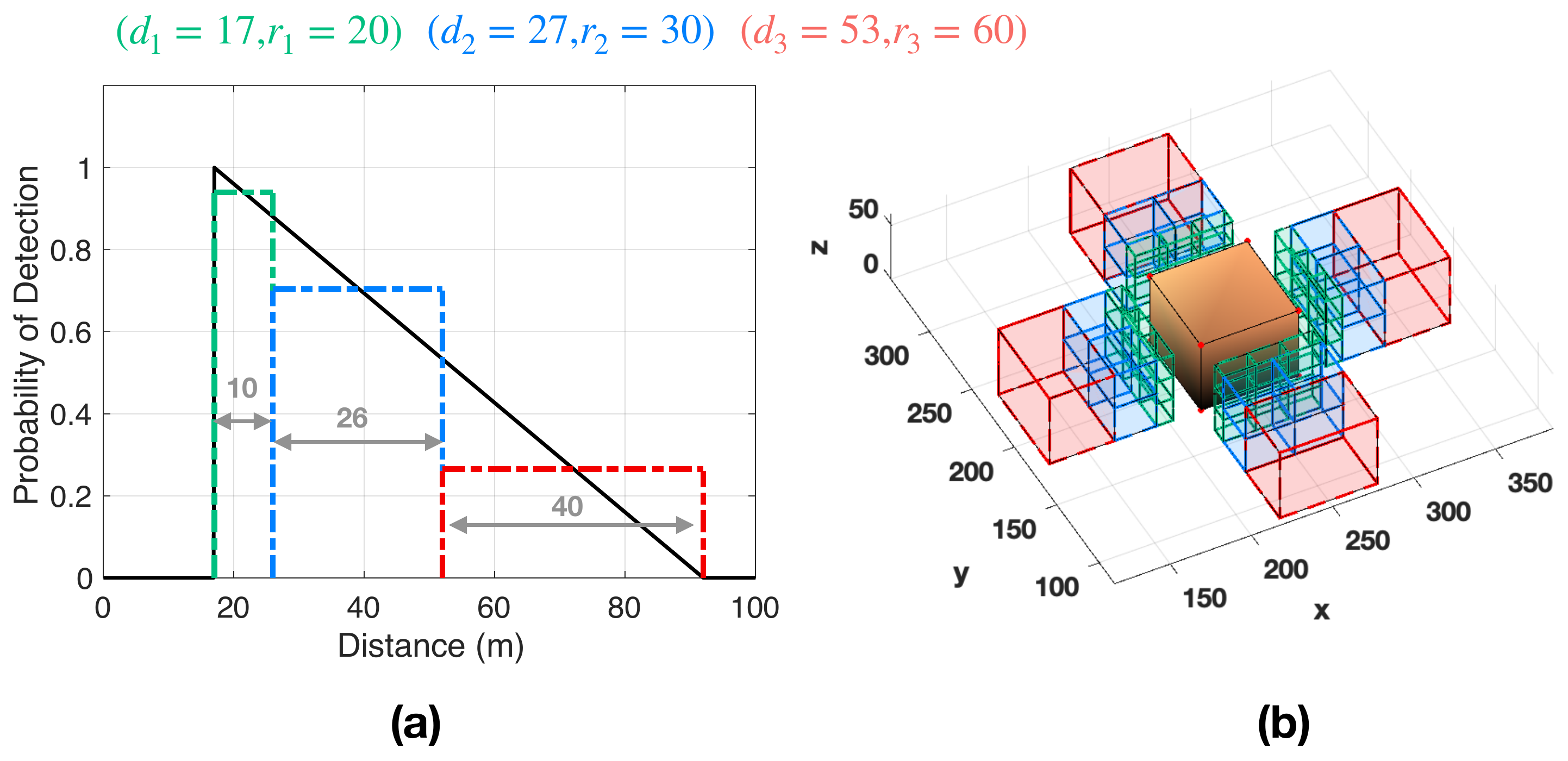}
	\caption{The figure illustrates the 3D Zone construction step used in our experiments. a) The probability of detection has been discretized into 3 zones, b) The area discretization around the object of interest and the formation of 3 zones (i.e., green, blue and red).}	
	\label{fig:res1_0}
	\vspace{-0mm}
\end{figure}

\subsection{Simulation Results}
\subsubsection{3D Search task} The first experiment aims to demonstrate the 3D search task discussed in Sec. \ref{ssec:3d_search_task} for a building represented as a cuboid with size 60m $\times$ 60m $\times$ 60m. The overall objective of the agent is to search around the building for survivors (i.e., search 4 faces of the building) with the required detection probability while at the same time optimize the mission objectives and at the end of the mission reach the goal region. The agent sensing model and in particular the detection probability used for this experiment is illustrated in Fig. \ref{fig:res1_0}(a). The figure shows a hypothetical discretization of the detection probability into $\mathcal{N}=3$ zones. For instance, the figure illustrates that the agent achieves a detection probability of 0.95 when it maintains a distance of $[17,27)$m from the object of interest (i.e., green zone) and a detection probability of 0.75 when the distance between the agent and the object of interest is in the range $[27,53)$m (i.e., blue zone). The agent's detection probability is zero for distances less than 17m and larger than 93m as shown in the figure. How the detection probability is discretized into various zones in order to capture the required detection levels is up to the designer and the mission objectives. For instance in this experiment we are interested in the first two zones i.e., green and blue zones, for detecting survivors with relatively high detection probabilities, whereas the red zone can be used for rapid spot checks and for mission assessment purposes.

\begin{figure*}
	\centering
	\includegraphics[width=\textwidth]{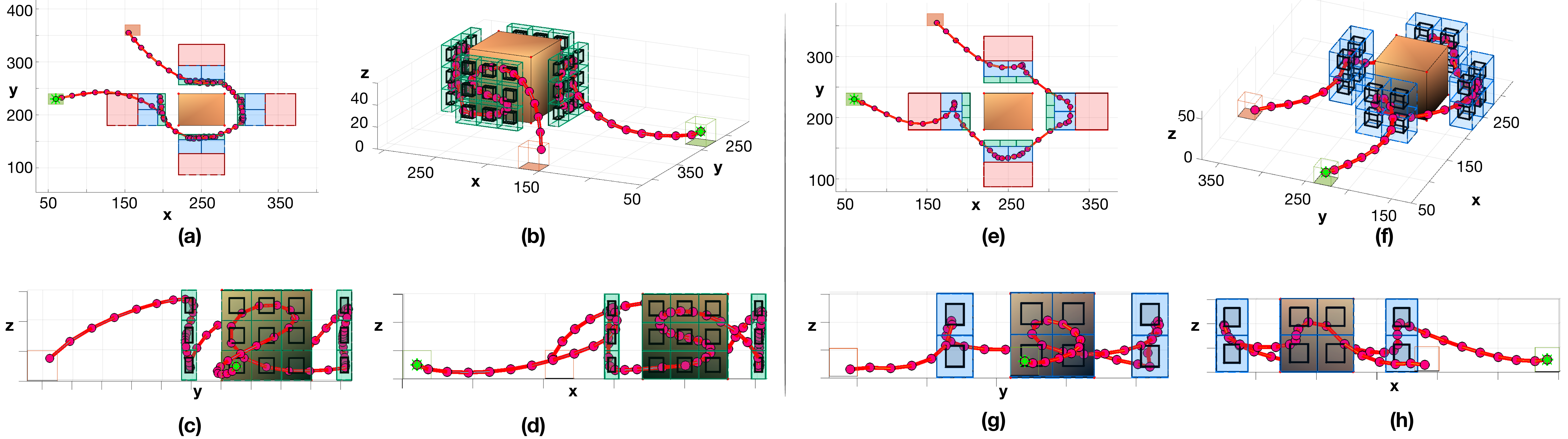}
	\caption{The figure illustrates the generated search plan for: (a)-(d) $\mathcal{Q}=0.9$ and (e)-(h) $\mathcal{Q}=0.7$ The agent passes through the appropriate zones and visits all the cuboids (and interior cubes) in its way in order to search each face of the object of interest with the desired detection probability.}	
	\label{fig:res1_12}
	\vspace{-0mm}
\end{figure*}

Once the various zones have been determined, the agent's FoV footprint can be computed for each zone according to Eqn. (\ref{eq:camera_model}). In particular the agent's FoV footprint is $r_1^2=20^2\text{m}^2$, $r_2^2=30^2\text{m}^2$ and $r_3^2=60^2\text{m}^2$ for zones 1,2 and 3 respectively (these quantities have been computed for distances  $d_1=17$m, $d_2=27$m and $d_3=53$m for zones 1,2 and 3 respectively). The generated 3D zones $\tilde{\mathcal{Z}}_1$, $\tilde{\mathcal{Z}}_2$ and $\tilde{\mathcal{Z}}_3$ have the following properties: a) $\tilde{\mathcal{Z}}_1$ has depth $D_1=10$m and detection probability $\hat{p_d}(\tilde{\mathcal{Z}}_1)=0.95$. Moreover, the length of zone 1 is $|\tilde{\mathcal{Z}}_1|=36$ i.e., $\tilde{\mathcal{Z}}_1$ is composed of 36 simple cuboids each of which has a size of $\sigma(Z_{1i}) = [20,20,10]$m, b) $\tilde{\mathcal{Z}}_2$ has depth $D_2=26$m, detection probability $\hat{p_d}(\tilde{\mathcal{Z}}_2)=0.75$ and the length $|\tilde{\mathcal{Z}}_2|=16$. Each cuboid of zone 2 has a size of $\sigma(Z_{2i}) = [30,30,26]$m, and finally c) for $\tilde{\mathcal{Z}}_3$, $D_3=40$, $\hat{p_d}(\tilde{\mathcal{Z}}_3)=0.25$, $|\tilde{\mathcal{Z}}_3|=4$ and each of the 4 simple cuboids of zone 3 has size of $\sigma(Z_{2i}) = [60,60,40]$m. To aid understanding, the whole concept is depicted in Fig. \ref{fig:res1_0}(a)-(b). Within each simple cuboid we create at its center an interior cube as shown in Fig. \ref{fig:res1_12}.

%\begin{figure}
%	\centering
%	\includegraphics[scale = 0.2]{figs/res1_2.pdf}
%	\caption{The figure illustrates the generated search plans for: (a)-(c) $\mathcal{Q}=0.9$ and (d)-(f) $\mathcal{Q}=0.7$.}	
%	\label{fig:res1_2}
%\end{figure}

In essence, the decomposition of the environment around the object of interest into the aforementioned 3D zones allows us to encode the agent sensing capabilities and the problem constraints in a form that can be easily incorporated into the proposed mathematical programming search framework. For instance, when the search task must be accomplished with a detection probability above 0.9, the agent will need to visit every one of the cuboids in Zone 1 in order to cover the whole object in 3D. On the other hand when the purpose is a quick and not so accurate search, the agent can perform the search task from a larger distance with larger FoV footprint which can be achieved through zone 2. These scenarios are illustrated in Fig. \ref{fig:res1_12}.

In particular Fig. \ref{fig:res1_12}(a)-(d) shows the generated search plan when the object of interest is required to be searched with a user defined detection probability of $\mathcal{Q}=0.9$. As we can observe from Fig. \ref{fig:res1_12}(a) (top-down view) in order to satisfy the required detection probability the agent passes through Zone 1 (i.e., green region). More specifically, as shown in Fig. \ref{fig:res1_12}(b) the agent departs from its initial location indicated by the green box and visits every cuboid (36 in total) of zone 1 in order to cover the whole surface of the object of interest. At the end of the mission the agent finishes inside the goal region indicated by the red box. For this experiment the mission objective i.e., Eqn. (\ref{eq:SAR_objective}) was used with $w_1=1$, $w_2=1$ and horizon time $T=90$ time-steps. Figure \ref{fig:res1_12}(c)-(d) shows in more detail the trajectory of the agent which was generated in order to accomplish the 3D search task. As shown, the agent maneuvers in such a way so that every simple cuboid inside Zone 1 is visited, thus the whole surface of the building is being searched.

Moreover, Fig. \ref{fig:res1_12}(e)-(h) shows the agent trajectory when we require a detection probability value $\mathcal{Q}=0.7$. To satisfy this constraint, the agent chooses to pass from Zone 2 (i.e., the blue region). Again, in this experiment we see that the agent visits all the cuboids of Zone 2 and finishes at the goal region as depicted more clearly in Fig. \ref{fig:res1_12}(f)-(h). Next, we take a closer look at the mission objective and how this affects the generated search plan.

\begin{figure*}
	\centering
	\includegraphics[width=\textwidth]{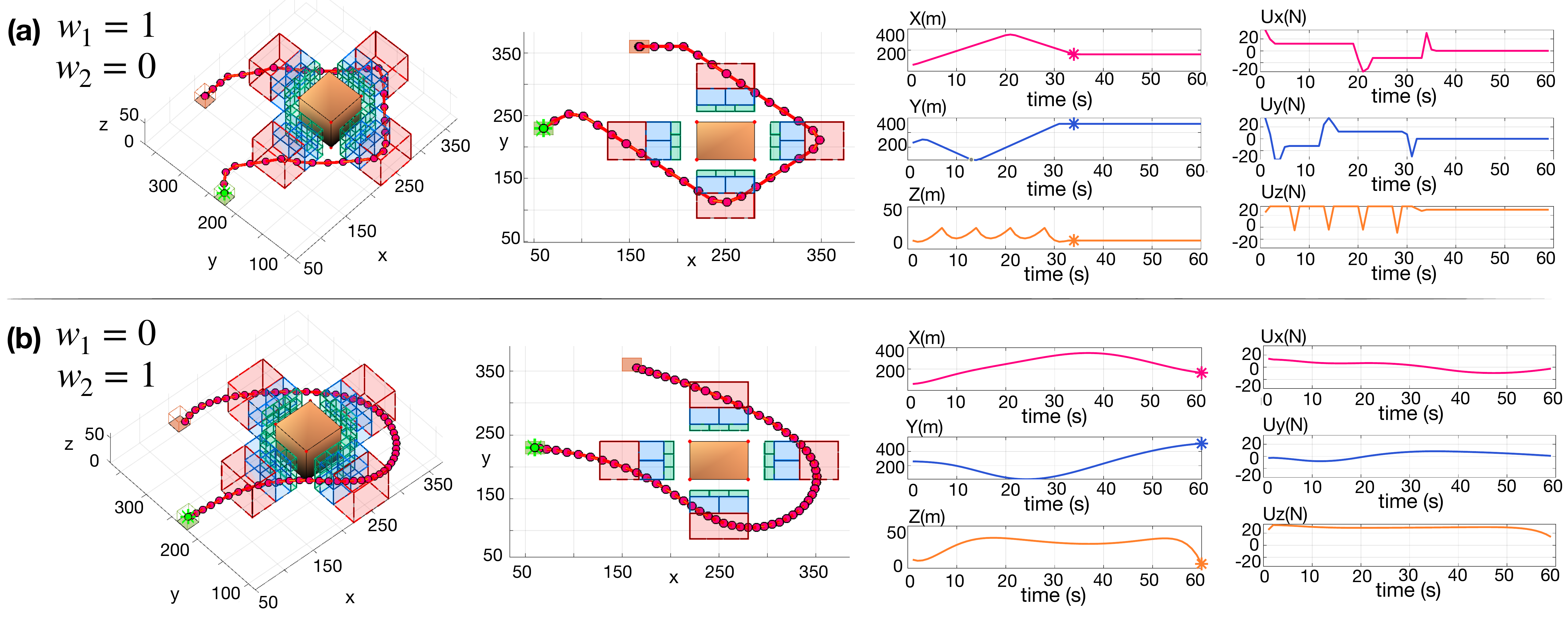}
	\caption{The figure illustrates the effect of the weights $w_1$ and $w_2$ of Eqn. (\ref{eq:SAR_objective}) on the generated search plan. (a) The generated search plan minimizes the mission execution time. (b) The generated plan optimizes the UAV energy efficiency by minimizing the deviations between consecutive control inputs.}	
	\label{fig:res2}
	\vspace{-0mm}
\end{figure*}

\subsubsection{Mission Objective} In this experiment we investigate how the parameters $w_1$ and $w_2$ of the objective function in Eqn. (\ref{eq:SAR_objective}) affect the generated search plan. As we have discussed in Sec. \ref{ssec:SARobj}, the mission objectives are linked to: a) the mission execution time and b) the UAV energy efficiency. In order to quantify the above objectives we have defined the Path Error Penalty (PEP) and the Input Fluctuation Penalty (IFP). The final mission objective in Eqn. (\ref{eq:SAR_objective}) is defined as the weighted combination of PEP and IFP and thus the parameters $w_1$ and $w_2$ determine the emphasis given to the two objectives. Figure \ref{fig:res2} shows a) the generated plan, b) the $(x,y,z)$ 3D coordinates of the agent over the horizon of length $T=60$ time-steps and c) the 3D control input $u_{1:T-1}$ for two configurations of the weights i.e., $(w_1=1,w_2=0)$ and $(w_1=0,w_2=1)$. In this experiment the object of interest and the 3D zones are as described in our previous experiment, and the agent passes through Zone 3. More specifically, Fig. \ref{fig:res2}(a) shows the generated search plan when only the mission execution time is considered. In this configuration the agent tries to reach the goal region as quickly as possible. This is achieved at time $\tau=34$ as indicated by the asterisk in the $(x,y,z)$ plot. Also, observe that the control inputs is spiky and not smooth which indicates rough changes in the control input and thus significant energy consumption. On the other hand, Fig. \ref{fig:res2}(b) shows the opposite effect when the weights are set as $(w_1=0,w_2=1)$. In particular in this scenario, only the IFP objective is active which minimizes the fluctuations between consecutive control inputs (and thus the energy consumption) also evident by the smoothness of the agent trajectory as shown in the figure. Also, observe that in this configuration the agent reaches the goal region at the end of the horizon at $\tau=T=60$. To conclude, depending on the mission requirements the proper weighting scheme can be applied in order to emphasize the attention given to the mission execution time and to the UAV energy efficiency.

\begin{figure*}
	\centering
	\includegraphics[width=\textwidth]{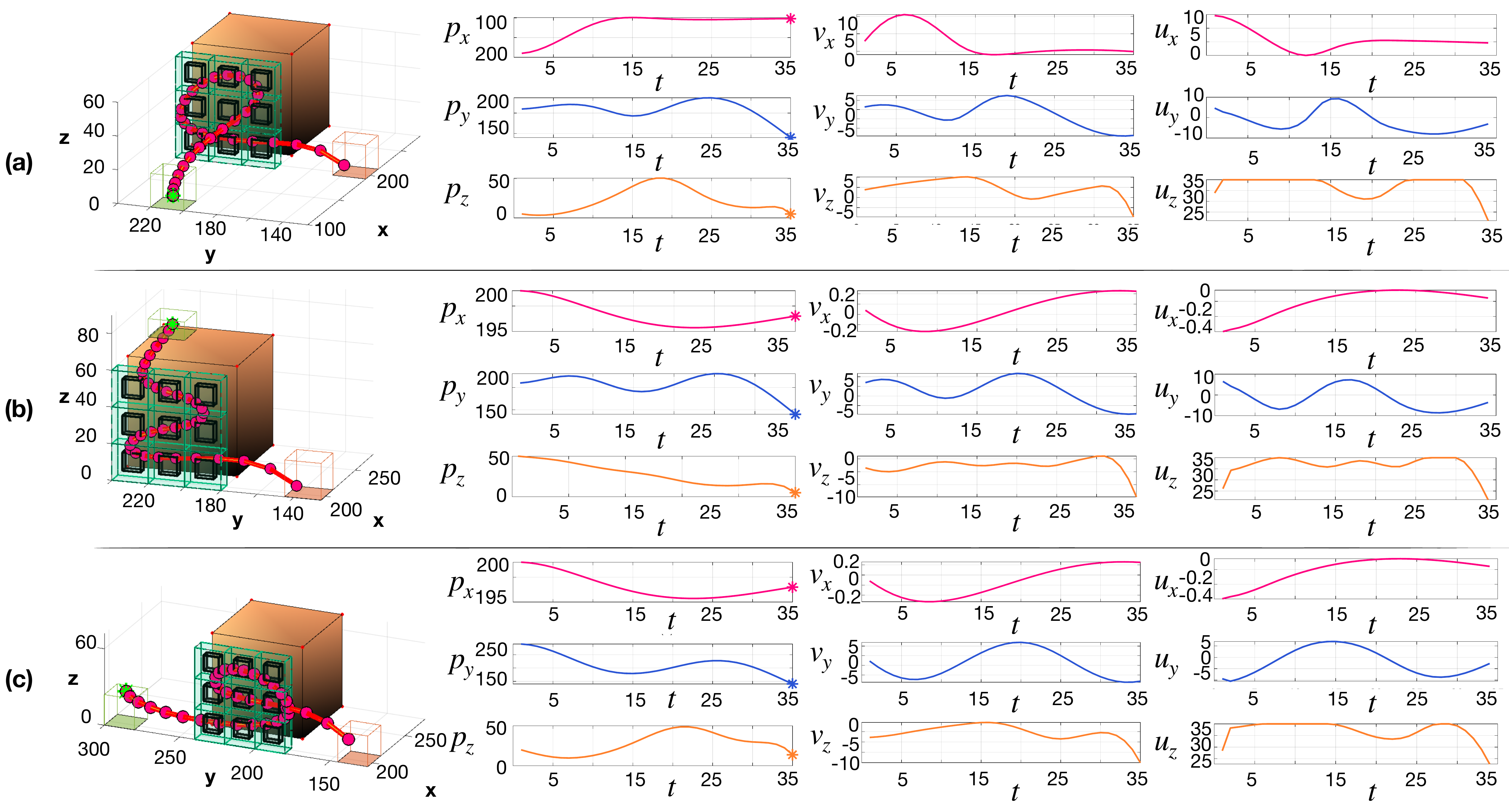}
	\caption{The figure illustrates the various search plans generated by the proposed framework for searching a single face of an object of interest. As it is shown, depending on the agent starting position the shape of generated trajectory changes in order to optimize the mission objective while incorporating the agent dynamics.}	
	\label{fig:res5}
	\vspace{-0mm}
\end{figure*}

Our next experiment illustrated in Fig. \ref{fig:res5} aims to investigate more closely the generated search plan and the behavior of the proposed framework. In this experiment we are interested in searching a single face of the object of interest as illustrated in the figure with a planning horizon of length $T=35$ and weights $(w_1=0,w_2=1)$. In order to get a better sense of the planning behavior, in this experiment we are varying the agent starting position i.e., in Fig. \ref{fig:res5}(a) the agent home depot is located in front of the object at a distance of 90m and the agent departs from the ground, whereas in Fig. \ref{fig:res5}(b) the agent is located directly above the object of interest at a distance of 85m from the ground. Finally, in Fig. \ref{fig:res5}(c) the agent departs from the side of the face to be searched at a distance of 50m. In each of the 3 experiments we monitor the agent 3D coordinates, velocity and input controls as shown in Fig. \ref{fig:res5}. As we can observe from the results shown, depending on the starting position, the UAV agent changes the way the simple cuboids inside the green zone are visited to produce the optimal search plan according to the mission objective and the UAV dynamics. In this experiment we are interested in minimizing the deviations between the consecutive control inputs as shown by the graphs in Fig. \ref{fig:res5}, which is achieved by incorporating the UAV dynamics into the planning process. This behavior would be more difficult to be achieved if we have decoupled the joint problem into two parts i.e., kinematic path planning and then adaptation to the UAV dynamics.

\subsubsection{Search planning in the presence of obstacles}
\begin{figure}
	\centering
	\includegraphics[width=\columnwidth]{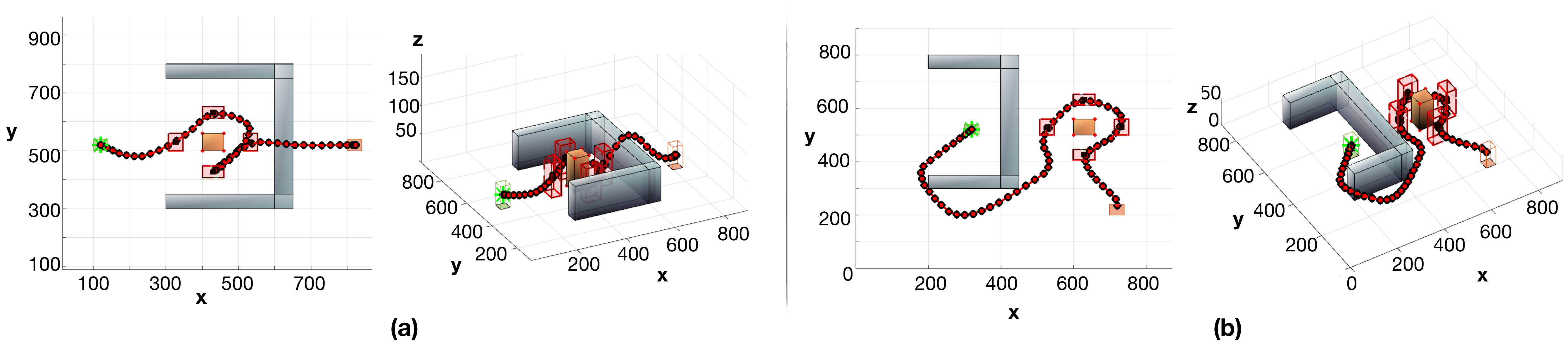}
	\caption{The figure illustrates the generated search plans in the presence of 3D obstacles.}	
	\label{fig:res3}
\end{figure}

Finally, the last experiment aims to investigate the behavior of the proposed search planning framework in the presence of obstacles. In this experiment we have set $(w_1=1,w_2=1)$, mission planning horizon $T=60$ time-steps and the specifications of the object and 3D zones are as previously. Figure \ref{fig:res3}(a)-(b) shows two scenarios where the agent needs to avoid the obstacle, search the object of interest and finish the mission inside the goal region. In the first scenario depicted in Fig. \ref{fig:res3}(a) the agent departs from its home depot (i.e., the green box), flies towards the object of interest, searches it in 3D and then proceeds to the goal region by flying above the obstacle. In Figure \ref{fig:res3}(b) the agent starts within the obstacle and goes around it in order to find and search the object of interest before moving inside the goal region. In this scenario we have restricted the maximum height that the agent can fly to 60m. Because the height of the obstacle is also 60m the agent has no longer the option to fly over the obstacle, thus in this scenario it goes around it as is illustrated in the figure.

Finally, we should mention that the computational complexity of the proposed search planning framework is mainly due to the 3D search task shown in Alg. \ref{alg:3d_search_task}. In particular, the main factor that drives the computational complexity is the number of binary variables which are required in mathematical program that implements the 3D search task. As the number of binary variables increases the produced search tree that is needed to be explored during the branch-and-bound/branch-and-cut \cite{Elf2001,MORRISON201679} optimization process increases in size and as a result more nodes are needed to be explored until the optimal solution is found. More specifically, the number of binary variables required in the 3D search task are equal to $\mathcal{B} = z_{tlci}+\tilde{z}_{tci}+\hat{z}_i, ~\forall~t,l,c,i$. Subsequently, the worst case scenario in terms of computational complexity is determined by the largest zone (i.e., the zone which contains the largest number of cuboids). In this case the total number of binary variables is equal to $\tilde{\mathcal{B}}=T |\tilde{\mathcal{Z}}| (L+1)$ where $T$ is the planning horizon, $|\tilde{\mathcal{Z}}|$ is zone length i.e., the number of cuboids contained inside zone $\tilde{\mathcal{Z}}$ and $L$ is the number of faces per cuboid. 

\begin{figure}
	\centering
	\includegraphics[width=\columnwidth]{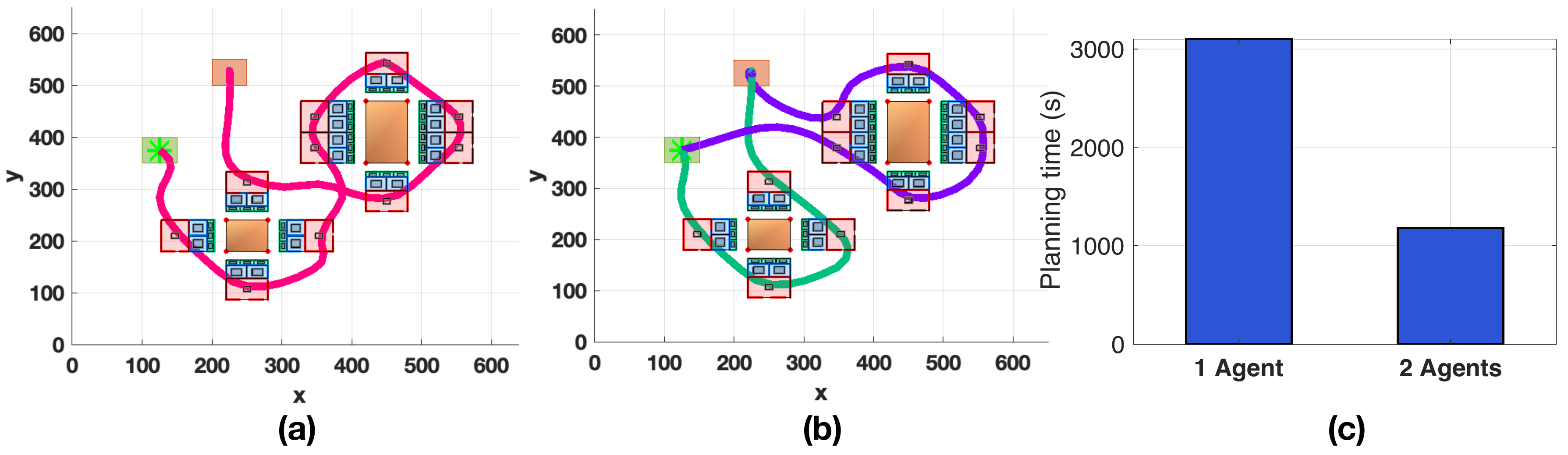}
	\caption{The figure illustrates the decomposition of the search planning problem into smaller sub-problems. Specifically, in this scenario 2 UAV agents are used to search 2 objects of interest (each agent is assigned to a different object). (a) The search planning conducted by 1 agent, (b) the search planning conducted by 2 agents, (c) the planning time for each scenario.}	
	\label{fig:res4_2}
	\vspace{-0mm}
\end{figure}

Although, the mixed integer quadratic (and linear) programs can be in general intractable for large problems \cite{Kannan1978}, recent advances in parallel branch-and-bound techniques \cite{Archibald2018,Munguia2019,Ralphs2018} allow moderate sizes of such problems to be solved exactly. Alternatively, adequate solutions can be obtained even for larger problems through various approximations and heuristics \cite{Lin2019,Nucamendi2016}. For instance a sub-optimal solution can be obtained by solving the problem in a rolling horizon fashion \cite{Wang2015} with reduced computational complexity. Another approach is to decompose the problem into small sub-problems and process each sub-problem sequentially or with multiple agents as depicted in Fig. \ref{fig:res4_2}. With the help of multiple agents, the planning horizon and number of cuboids that need to be visited by each agent is reduced and so are the number of binary variables required for implementing the search task.

In this section we have conducted a thorough synthetic evaluation analysis of the proposed approach, demonstrating its effectiveness for different parameter configurations. Specifically, in Sec. 6.2.1 we have demonstrated the behavior of the system for different detection probability values, in Sec. 6.2.2 we have shown the effect of the weights $w_1$ and $w_2$ on the mission's execution time and on the trajectory of the UAV and finally, in Sec. 6.2.3 we have shown the behavior of the system in the presence of obstacles, and we have discussed its computational complexity. 

 Overall, the proposed framework aims to augment the traditional SAR missions with precise and efficient automated search capabilities and with improved organization and planning, by automating the trajectory planning process and guidance of an autonomous UAV agent. Manually, operating a UAV in SAR missions is an error-prone process and requires a high degree of human expertise. Moreover, optimally guiding the UAV agent by manual control is very challenging. Precise navigation and time-efficient searching can save lives during SAR missions which is the main motivation for this work. The proposed framework can be used alongside the SAR mission, to optimally plan search operations in areas which are inaccessible to the rescue crew and in scenarios where the infrastructure is destroyed or disrupted. 

Future work, will investigate the real-world implementation of the proposed system, and its integration into our existing multi-drone tasking platform \cite{terzi2019swifters,Terzi2}. In particular we will investigate how the generated trajectories can be translated into low-level control inputs which can be executed by the UAV's on-board controller \cite{Luis2020,Yang2008}. Additionally, we aim to investigate what are the benefits of using the proposed automated framework in real-world SAR scenarios, and how it compares to the traditional SAR missions. In particular, we plan to conduct real field tests, in collaboration with the Cyprus Civil Defense (CCD) \cite{CCD}, in order to gain key insights on the real-world performance of the system. Moreover, the experimental testing of the proposed framework in the field will allow us to assess its applicability in real-world conditions (e.g., in various environmental conditions) and identify its limitations i.e., modeling inefficiencies and hardware/platform limitations.

\section{Conclusion} \label{sec:conclusion}

In this work we have proposed a novel search planning framework which automates the UAV-based search missions in 3D environments. The proposed framework generates search plans which allow the UAV agent to search all the faces of an object of interest with the desired probability of detection while at the same time avoiding collisions with the obstacles in the environment. 
To enable the 3D search task we decompose the environment around the object of interest into simple cuboids and we form 3D zones which are then encoded into constraints using mathematical programming techniques. Finally, the 3D search constraints along with the low-level mission constrains such as the UAV dynamical and sensing model and the mission objectives are combined to form a mixed integer program which is then solved exactly using off-the-shelf MIQP solvers. 

\section{Acknowledgements}
This work is supported by the European Union's Horizon 2020 research and innovation programme under grant agreement No 739551 (KIOS CoE), by the European Union Civil Protection Call for proposals UCPM-2019-PP-AG grant agreement No 873240 (AIDERS) and from the Republic of Cyprus through the Directorate General for European Programmes, Coordination and Development.
\balance
\bibliographystyle{IEEEtran}
\bibliography{main}

\end{document}